\documentclass[10pt,letterpaper]{article}

\usepackage[T1]{fontenc}
\usepackage{lmodern}
\usepackage{newtxtext}
\usepackage[english]{babel}
\usepackage[top=2.25cm,bottom=2.5cm,left=2.5cm,right=2.5cm]{geometry}
\usepackage{microtype}
\usepackage{parskip}
\usepackage{placeins}
\usepackage{needspace}
\usepackage{setspace}
\usepackage{graphicx}
\usepackage{wrapfig}
\usepackage{amsmath,amssymb,bm}
\usepackage{booktabs,tabularx,multirow}
\usepackage[dvipsnames,table]{xcolor}
\usepackage[most]{tcolorbox}
\usepackage{titlesec}
\usepackage{caption,subcaption}
\usepackage{algorithm}
\usepackage{algpseudocode}
\usepackage{xspace}
\usepackage{etoolbox}
\usepackage[numbers,sort&compress]{natbib}
\usepackage{hyperref}
\usepackage[noabbrev,nameinlink]{cleveref}
\usepackage{amsthm}
\newtheorem{proposition}{Proposition}

\definecolor{PaperAccent}{HTML}{FC8066}
\definecolor{PaperInk}{HTML}{213039}
\definecolor{PaperCream}{HTML}{FEECC3}
\definecolor{MethodAccent}{HTML}{EC866D}
\definecolor{NoteInk}{HTML}{59636D}
\definecolor{PanelGray}{HTML}{F3F4F5}

\DeclareTextFontCommand{\textbf}{\bfseries\sffamily}
\titleformat*{\section}{\Large\sffamily\bfseries}
\titleformat*{\subsection}{\large\sffamily\bfseries}
\titleformat*{\subsubsection}{\normalsize\sffamily\bfseries}
\titleformat*{\paragraph}{\itshape}
\titlespacing*{\paragraph}{0pt}{1.5ex plus 0.3ex minus 0.2ex}{0.8em}
\DeclareCaptionLabelSeparator{papersep}{}
\DeclareCaptionFormat{papercaption}{{\sffamily\bfseries #1 #2} #3}
\hypersetup{colorlinks=true,linkcolor=PaperAccent,citecolor=PaperAccent,urlcolor=PaperAccent}
\newcommand{\para}[1]{\paragraph{\textbf{\boldmath #1}}}
\newcommand{\papername}{Loo-P}
\newcommand{\papersubtitle}{Scaling Looped Transformers with Structured Residuals}
\newcommand{\paperauthors}{}
\newcommand{\paperaffiliations}{}
\renewcommand{\author}[2][]{\renewcommand{\paperauthors}{#2}}
\newcommand{\affiliation}[2][]{\renewcommand{\paperaffiliations}{#2}}
\author[]{Yiran Xu\textsuperscript{1}, Yawen Luo\textsuperscript{1,2,*}, Hanwen Jiang\textsuperscript{1}, Chongjian Ge\textsuperscript{1}, Zefan Cai\textsuperscript{1,3,*}}
\affiliation[]{\small\textsuperscript{1}Adobe Research\\
  \textsuperscript{2}Chinese University of Hong Kong\quad
  \textsuperscript{3}University of Wisconsin--Madison}
\newcommand{\papercontributions}{}
\newcommand{\paperdate}{}
\newcommand{\paperabstract}{%
Looped Transformers trade independent parameters for repeated computation, yet effective training, scaling, and performance at matched compute remain open questions.
We introduce \ours{}, a pretraining recipe that combines structured residuals with hyperparameter transfer.
Normalized module-wise and visit-wise coefficients allocate computation across attention, feed-forward modules, and recurrent visits, while coordinated residual and optimizer scaling support transfer across recurrence counts and unique depths.
Selected on small task-specific proxies, these coefficients add no learned parameters.
In language pretraining, \ours{} achieves lower validation cross-entropy than prior looped baselines while retaining effective learning rates as recurrence and unique depth grow, separately or together.
Beyond scalability, our evaluation across language pretraining, 3D reconstruction, class-to-image generation, and text-to-image generation reveals that \textbf{the effectiveness of parameter sharing is task-dependent}.
Under each task's compute-matching protocol, sharing offers a more favorable parameter--performance trade-off in the studied vision tasks than in language pretraining, where \ours{} narrows but does not close the gap to independent layers.
In 3D reconstruction, \ours{} improves both pose accuracy and novel-view quality with \textbf{34.6\% fewer parameters} at matched encoder compute.
Image generative models retain comparable pretraining losses, and our text-to-image model surpasses a non-looped baseline on GenEval and DPG-Bench after fine-tuning with \textbf{45\% fewer parameters} at matched compute.
This contrast motivates the hypothesis that sharing is particularly effective when successive stages can reuse a common refinement operation, whereas other tasks benefit more from distinct transformations.
Our results connect scalable training with where repeated computation can substitute for independent parameters.
}
\hypersetup{pdftitle={Loo-P beyond Language: Scaling Looped Transformers with Structured Residuals},pdfauthor={Yiran Xu, Yawen Luo, Hanwen Jiang, Chongjian Ge, Zefan Cai},pdfsubject={}}

\newcommand{\makepapertitle}{%
  \begin{tcolorbox}[
    enhanced,frame hidden,colback=PaperCream,arc=10pt,
    left=0.5cm,right=0.5cm,top=0.5cm,bottom=0.5cm,
    before skip=0pt,after skip=0.3cm,
    grow to left by=1.5pt,grow to right by=1.5pt]
    \raggedright
    {\huge\sffamily\bfseries
      \textcolor{MethodAccent}{\papername} beyond Language\ifdefempty{\papersubtitle}{}{: \papersubtitle}\par}
    \ifdefempty{\paperauthors}{}{\medskip{\sffamily\bfseries\paperauthors\par}}
    \ifdefempty{\paperaffiliations}{}{\smallskip{\paperaffiliations\par}}
    \ifdefempty{\papercontributions}{}{\smallskip{\small\papercontributions\par}}
    \medskip
    {\color{PaperInk}\textbf{Abstract}\par\smallskip\paperabstract\par}
    \ifdefempty{\paperdate}{}{\medskip{\small\textbf{Date:} \paperdate\par}}
  \end{tcolorbox}%
}

\newcommand{\ours}{\textcolor{FutureOrange}{\textbf{Loo-P}}}
\DeclareRobustCommand{\ours}{\texorpdfstring{\textcolor{MethodAccent}{\textbf{Loo-P}}}{Loo-P}\xspace}

\begin{document}
\makepapertitle
\begingroup
\renewcommand{\thefootnote}{\fnsymbol{footnote}}
\footnotetext[1]{Work done as an intern at Adobe Research.}
\endgroup
\DeclareRobustCommand{\teaserbaselineglyph}{%
  \raisebox{0.45ex}{\textcolor{NoteInk}{%
    \rule{0.42em}{0.55pt}\kern0.16em%
    \rule{0.42em}{0.55pt}\kern0.16em%
    \rule{0.42em}{0.55pt}}}%
}
\begin{figure}[!htbp]
  \centering
  \captionsetup[subfigure]{skip=2pt}
  \begin{subfigure}[t]{0.42\textwidth}
    \centering
    \includegraphics[width=\linewidth]{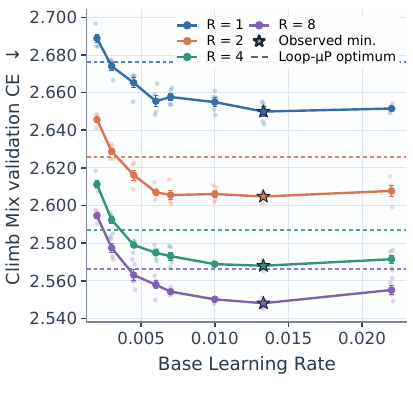}
    \caption{How to scale}
    \label{fig:teaser-hp-transfer}
  \end{subfigure}\hfill
  \begin{subfigure}[t]{0.56\textwidth}
    \centering
    \includegraphics[width=\linewidth]{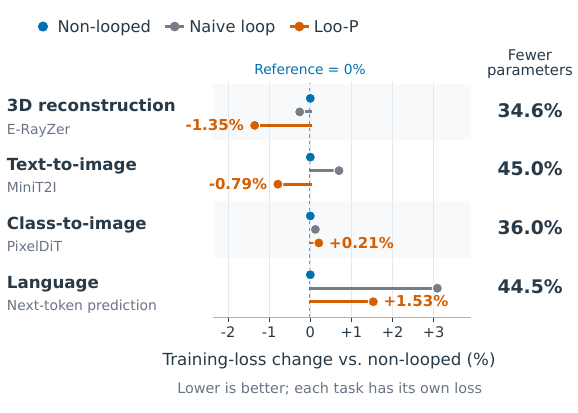}
    \caption{Where sharing works}
    \label{fig:teaser-sharing}
  \end{subfigure}

  \vspace{1pt}
    \begin{subfigure}[t]{0.42\textwidth}
      \centering
      \includegraphics[width=\linewidth]{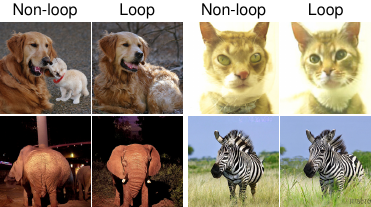}
      \caption{Image generation}
      \label{fig:teaser-middle}
    \end{subfigure}\hfill
    \begin{subfigure}[t]{0.56\textwidth}
      \centering
      \includegraphics[width=\linewidth]{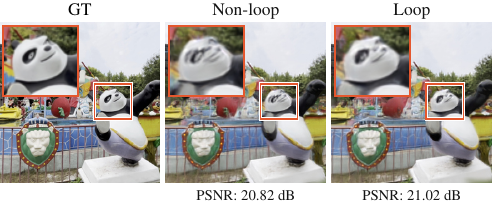}
      \caption{Novel view synthesis}
      \label{fig:teaser-right}
    \end{subfigure}
  \caption{\textbf{How to scale Looped Transformers, and where parameter sharing works.}
  (\textbf{a}) \ours shows stable HP transfer across the loop-axis $R$ while achieving lower validation CE than our Loop-$\mu$P~\cite{wang2026residual} reproduction (dashed lines, \teaserbaselineglyph).
  (\textbf{b}) Vision tasks prefer weight sharing. At matched compute within each task, the final $20\%$ window training-loss change relative to the non-looped model is shown alongside the percentage of parameters saved by sharing.
  \ours retains comparable or lower loss in the vision tasks, but a gap remains in language.
  (\textbf{c}) Paired PixelDiT~\cite{yu2026pixeldit} samples on ImageNet-256~\cite{deng2009imagenet}. Our looped model shows comparable or even better quality.
  (\textbf{d}) E-RayZer~\cite{zhao2026erayzer} renderings show the same target view, with GT, non-looped, and looped outputs from left to right. The looped model shows better PSNR.
  }
  \label{fig:teaser}
\end{figure}

\section{Introduction}
\label{sec:intro}




Looped Transformers increase effective depth by repeatedly applying a shared set of layers, allowing a compact model to perform more computation without adding new parameters at every depth.
This idea builds on depth-wise recurrence in Universal Transformers~\cite{dehghani2018universal} and cross-layer parameter sharing in ALBERT~\cite{lan2019albert}.
Recent work has extended shared computation to recurrent-depth language models~\cite{geiping2025huginn} and visual generation~\cite{goyal2026elt,chng2026loopeddit,derakhshani2026lift}, highlighting its potential for parameter efficiency and flexible computation.

The central challenge is to make this parameter efficiency competitive with non-looped models at the same compute budget.
A non-looped Transformer spends its computation on distinct learned transformations, whereas a looped model must make repeated use of the same weights.
Matching effective depth or FLOPs therefore does not guarantee comparable performance.
Methods such as Mixture-of-Recursions~\cite{bae2025mor} and Loopie~\cite{gao2026loopie} report encouraging results through adaptive computation and shared experts, respectively, but these gains depend on the accompanying modeling and training choices.
Even with effective training, the value of sharing may depend on whether a task benefits more from reusing a transformation or learning distinct transformations at successive depths.
This leads to our central question: \emph{How can we effectively train and scale Looped Transformers, and to what extent can repeated computation substitute for independent parameters across different tasks?}

We address training and scalability through two complementary requirements: improving what a shared Transformer block learns and enabling hyperparameter (HP) transfer as its shape changes.
For the first, we study how residual computation is allocated across modules and visits.
Attention and feed-forward modules perform different transformations, while repeated visits expose the same weights to evolving representations.
We introduce \emph{module-wise} coefficients to balance attention and feed-forward residuals, and \emph{visit-wise} coefficients to vary their contribution along the recurrent trajectory.
These coefficients provide task-specific control over shared computation without introducing new learned weights for each visit.
Their normalization separates this allocation from the global residual scale, allowing us to improve loss while retaining a consistent scaling recipe.

For the second requirement, a good configuration should remain useful beyond the small proxy model on which it was selected.
Increasing recurrence or unique depth changes the computation and the gradients accumulated by shared parameters, so effective hyperparameters need not transfer automatically.
Building on residual parameterization and HP-transfer studies~\cite{yang2024tensor,dey2025completep,wang2026residual}, we pair the module-wise and visit-wise coefficients with global residual and optimizer scaling rules.
A key difference from Loop-$\mu$P is how Core residuals scale with unique depth $L$: we adopt inverse-depth scaling, with exponent $\alpha_L=1$, rather than inverse-square-root scaling, with $\alpha_L=\frac{1}{2}$.
Following the depth-scaling principle of CompleteP~\cite{dey2025completep}, this choice, combined with our residual allocation and optimizer scaling, supports more consistent learning-rate transfer in our experiments.

Together, these components form \ours{}, a pretraining recipe for Looped Transformers that combines structured residual design with hyperparameter transfer.
We use language pretraining as the primary testbed, varying recurrence, unique depth, and both together to examine whether a reference learning rate remains effective as the model scales.
In the reported transfer sweeps, \ours{} combines aligned learning-rate minima with lower validation cross-entropy than our sandwich adaptation of Loop-$\mu$P~\cite{wang2026residual}, as illustrated in Figure~\ref{fig:teaser-hp-transfer}.
The recipe thus improves learning at a given model shape while supporting reuse of effective settings as that shape grows.

The matched-compute language comparison also exposes a remaining limitation: \ours{} narrows the gap between naive looping and independent layers, but does not close it.
Is this sharing penalty intrinsic to looping, or does it depend on the modality?
To investigate, we apply the same residual-design framework to 3D reconstruction, class-to-image generation (C2I), and text-to-image generation (T2I), comparing looped and non-looped models under each task's compute accounting.
This extends the evaluation beyond whether the recipe scales to whether the resulting parameter savings preserve task performance.

Our cross-task evaluation reveals a more favorable parameter--performance trade-off in the studied vision tasks than in language pretraining (Figure~\ref{fig:teaser-sharing}).
In E-RayZer~\cite{zhao2026erayzer}, \ours{} improves pose accuracy and novel-view PSNR with $34.6\%$ fewer parameters at matched encoder compute.
The image-generation models retain comparable pretraining losses; after fine-tuning, the T2I model surpasses its non-looped counterpart on GenEval and DPG-Bench with $45\%$ fewer denoiser parameters at matched nominal denoiser compute.
This does not mean that sharing improves every metric: some distributional quality measures still favor the non-looped models.
Rather, the contrast with language shows that the cost of replacing independent parameters with repeated computation is task-dependent.

This empirical finding motivates a scientific hypothesis: sharing may work particularly well when successive stages can reuse a common refinement operation, while tasks requiring more distinct transformations may benefit more from independent layers.
Multi-view reconstruction and conditional generation provide settings in which shared computation could repeatedly refine an evolving representation.
Our comparisons establish the task-dependent trade-off, but do not isolate its cause, since architectures, data, and objectives also differ across tasks.
By bringing this question together with a practical training recipe, \ours{} provides both a method for scaling Looped Transformers and evidence about where their parameter efficiency is most useful.

\section{Method}
\label{sec:method}
Our goal is to improve vanilla Looped Transformers while enabling effective hyperparameter (HP) transfer.
We first observe that attention and MLP modules play distinct roles in looped layers, motivating a module-wise residual strategy (Sec.~\ref{sec:module-residuals}) that scales their residual branches differently. 
Furthermore, because representations evolve across visits in a loop, a visit-wise residual strategy (Sec.~\ref{sec:visit-residuals}) further improves performance. 
Finally, we unify these factors into a coherent residual parameterization, \ours, to preserve the residual scale and support reliable HP transfer (Sec.~\ref{sec:hp-transfer}).

\subsection{Sandwich topology}
\label{sec:topology}
We use a sandwich architecture~\cite{geiping2025huginn} with a Prelude of $P$ blocks, a Core of $L$ blocks, and a Coda of $C$ blocks.
The Core is applied $R$ times with shared weights, while the Prelude and Coda are each applied once.
The model therefore has $P+L+C$ distinct blocks and an effective depth of $P+RL+C$, counting each Core block once per visit.

Writing $\mathcal{P}_{P}$ and $\mathcal{C}_{C}$ for the Prelude and Coda stacks, and $\mathcal{F}_{L}^{(r)}$ for the $r$-th visit to the $L$-block Core, the full forward computation is
\begin{equation}
  f_{P,L,R,C}(x)
  = \bigl(
    \mathcal{C}_{C}
    \circ\underbrace{\mathcal{F}_{L}^{(R-1)}\circ\cdots\circ\mathcal{F}_{L}^{(0)}}_{R\text{ visits, shared weights}}
    \circ\mathcal{P}_{P}
    \bigr)(x).
  \label{eq:topology-forward}
\end{equation}
Here we omit the input read-in and output read-out layers (\textit{e.g.,} embedding and unembedding layers in a language model). 
All $\mathcal{F}_{L}^{(r)}$ share the same learned Core weights.
We use this topology for all our models across different modalities.

Specifically, for Core block $\ell \in \{0,\ldots,L-1\}$ on visit $r \in \{0,\ldots,R-1\}$ with input $x_{r,\ell}$ (where $x_{r+1,0}=x_{r,L}$), we write
\begin{align}
  z_{r,\ell} &= x_{r,\ell}
    + \gamma_{A,r}\,\mathrm{Attn}_{\ell}\!\left(\mathrm{Norm}(x_{r,\ell})\right), \label{eq:core-attn}\\
  x_{r,\ell+1} &= z_{r,\ell}
    + \gamma_{F,r}\,\mathrm{FFN}_{\ell}\!\left(\mathrm{Norm}(z_{r,\ell})\right), \label{eq:core-ffn}
\end{align}
where $\mathrm{Attn}$ denotes attention, $\mathrm{FFN}$ denotes feed-forward, and $\mathrm{Norm}$ denotes pre-normalization (RMSNorm in our language models). The Prelude and Coda retain unit residual coefficients.
We factorize the Core branch coefficients into three interpretable components:
\begin{equation}
  \gamma_{A,r} = g_{R,L}\,w_A\,v_r,
  \qquad
  \gamma_{F,r} = g_{R,L}\,w_F\,v_r,
  \label{eq:residual-factorization}
\end{equation}
where $(w_A, w_F)$ balances the attention and $\mathrm{FFN}$ branches, $v_r$ modulates the contribution across loop visits, and $g_{R,L}$ governs the global residual scale across loop count $R$ and Core depth $L$.

\subsection{Module-wise Residuals}
\label{sec:module-residuals}

Attention and $\mathrm{FFN}$ modules perform fundamentally different transformations even at the same depth~\cite{team2026attention}.
To adjust their relative residual contributions while preserving the squared coefficient budget, we parameterize $w_A$ and $w_F$ via a positive ratio $\rho > 0$:
\begin{equation}
  w_A = \rho\sqrt{\frac{2}{1+\rho^2}},
  \qquad
  w_F = \sqrt{\frac{2}{1+\rho^2}}.
  \label{eq:module-factors}
\end{equation}
These factors satisfy two properties:
\begin{equation}
  \frac{w_A}{w_F} = \rho,
  \qquad
  w_A^2 + w_F^2 = 2.
  \label{eq:module-conservation}
\end{equation}
For any scan-selected $\rho > 0$, the squared coefficients sum to $2$ (matching the standard unit-coefficient baseline $w_A = w_F = 1$).
Thus, $\rho$ reallocates residual capacity between cross-token routing ($\mathrm{Attn}$) and token-wise transformation ($\mathrm{FFN}$) while keeping the block-level squared residual scale invariant.
(Note that this invariance holds at the coefficient level and does not assume identical activation norms across modules.)

\subsection{Visit-wise Residuals}
\label{sec:visit-residuals}

Although Core weights are shared, hidden representations evolve from raw Prelude features toward refined predictions over successive visits.
Applying identical step sizes across all visits can over-perturb early, unaligned states or under-refine later iterations.
We therefore modulate the residual contribution along the visit index $r \in \{0,\ldots,R-1\}$ with a cosine schedule:
\begin{equation}
  v_r = 1 + \beta\cos\!\left(\frac{\pi(r+\frac{1}{2})}{R}\right),
  \qquad |\beta| < 1.
  \label{eq:visit-profile}
\end{equation}
For $|\beta|<1$, every multiplier $v_r$ remains strictly positive, and the profile satisfies two conservation properties:
\begin{equation}
  \frac{1}{R}\sum_{r=0}^{R-1} v_r = 1,
  \qquad
  v_0 = 1 \;\;\text{when } R=1.
  \label{eq:visit-conservation}
\end{equation}
Choosing $\beta < 0$ assigns smaller multipliers to early visits and larger multipliers to later visits, redistributing residual updates along the recurrent trajectory without changing the average multiplier across visits.

Crucially, these two normalization constraints decouple
residual allocation from model-shape-dependent scaling.
For fixed $\rho>0$ and $|\beta|<1$, both the module-wise
and visit-wise factors remain bounded independently of
$R$ and $L$, preserving the underlying scaling exponents
of $g_{R,L}$.

To illustrate this property, consider an idealized setting
where the attention and FFN updates, $\Delta_A$ and
$\Delta_F$, are zero-mean, uncorrelated, and have equal
second moments $\sigma^2$. The nominal squared residual
contribution of a Core block on visit $r$ is
\begin{align}
  \mathbb{E}\left[
    \left\|\gamma_{A,r}\Delta_A
    +\gamma_{F,r}\Delta_F\right\|^2
  \right]
  &\approx
  g_{R,L}^2v_r^2(w_A^2+w_F^2)\sigma^2
  \nonumber\\
  &=2g_{R,L}^2v_r^2\sigma^2.
  \label{eq:block-variance-invariance}
\end{align}
Furthermore, the visit profile satisfies
\begin{equation}
  \frac{1}{R}\sum_{r=0}^{R-1}v_r=1,
  \qquad
  \frac{1}{R}\sum_{r=0}^{R-1}v_r^2=
  \begin{cases}
    1, & R=1,\\
    1+\beta^2/2, & R\geq 2.
  \end{cases}
  \label{eq:visit-moments}
\end{equation}
Thus, module-wise allocation preserves the nominal
squared coefficient budget, while visit-wise allocation
preserves the mean multiplier and introduces only a
bounded second-moment factor. Neither changes the
depth- or recurrence-scaling exponents of the residual
coefficients.

We select $(\rho,\beta)$ through task-specific proxy
scans and hold them fixed during transfer across model
shapes. Both factors introduce no additional learned
parameters or visit-specific weight copies.

Appendix~\ref{app:theory} proves these coefficient identities and distinguishes them from assumptions about residual covariance and optimization.

\subsection{\ours: Hyperparameter Transfer}
\label{sec:hp-transfer}

To support HP transfer across recurrence counts $R$ and unique Core depths $L$, we pair the residual factors from Secs.~\ref{sec:topology}--\ref{sec:visit-residuals} with a global Core scaling factor rule written in normalized depth and recurrence ratios:
\begin{equation}
  g_{R,L} =  m_R^{-\alpha_R}\, m_L^{-\alpha_L},
  \qquad m_R = \frac{R}{R_0},
  \qquad m_L = \frac{L}{L_0},
  \label{eq:global-residual}
\end{equation}
where $R_0$ and $L_0$ denote reference recurrence count and unique Core depth, and $(\alpha_R, \alpha_L)$ control how the residual scale adapts along the recurrence and unique-depth axes.

\para{Scaling exponents in $\alpha \in [\frac{1}{2}, 1]$ and the recurrence $R$-axis.} In scaling and hyperparameter-transfer literature, exponents along depth or recurrence axes are commonly examined within $\alpha \in [\frac{1}{2}, 1]$, spanning uncorrelated accumulation ($\alpha = \frac{1}{2}$), 
or stronger coherent scaling ($\alpha = 1$).
Prior works adopt different instantiations along this range: Depth-$\mu$P~\cite{yang2024tensor} establishes $L^{-\frac{1}{2}}$ ($\alpha_L = \frac{1}{2}$) depth scaling to preserve feature learning in deep residual networks; 
CompleteP~\cite{dey2025completep} examines depth scaling with $\alpha_L \in \{\frac{1}{2}, 1\}$ alongside width factors, but finally uses $\alpha_L = 1$ to reduce the laziness during feature learning; 
Loop-$\mu$P~\cite{wang2026residual} scales loop count linearly ($\alpha_R = 1$) for weight-tied recurrence while retaining $\alpha_L = \frac{1}{2}$ along unique depth. 
Along the recurrence axis $R$, shared weights can correlate updates across visits, so independence across visits cannot be assumed.
Scaling by $R^{-1}$ keeps the sum of residual coefficients across visits fixed at each unique layer.
With this residual rule held fixed, our learning rate fit (Fig.~\ref{fig:r-axis-lr-coordinate}, Appendix~\ref{app:recurrence-lr-analysis}) supports recurrence-independent block LRs.
We therefore follow Loop-$\mu$P~\cite{wang2026residual} and use $\alpha_R = 1$ in Eq.~\ref{eq:global-residual}.

\para{Unique-depth $L$ scaling.}
Along unique Core depth $L$, our language recipe uses $\alpha_L=1$, whereas Loop-$\mu$P~\cite{wang2026residual} uses $\alpha_L=\frac{1}{2}$.
Repeated visits couple the computations within the Core, motivating us to examine scaling beyond independent residual accumulation.
Under an explicit positive-drift assumption, the scalar example in Appendix~\ref{app:theory} shows that $\alpha_L=1$ yields a bounded forward amplification with a finite large-depth limit, while $\alpha_L=\frac{1}{2}$ permits amplification to grow.
This motivates stronger depth control; the resulting HP transfer is assessed empirically in Figure~\ref{fig:depth-transfer}.

\begin{figure}[!htbp]
  \centering
  \includegraphics[width=\linewidth]{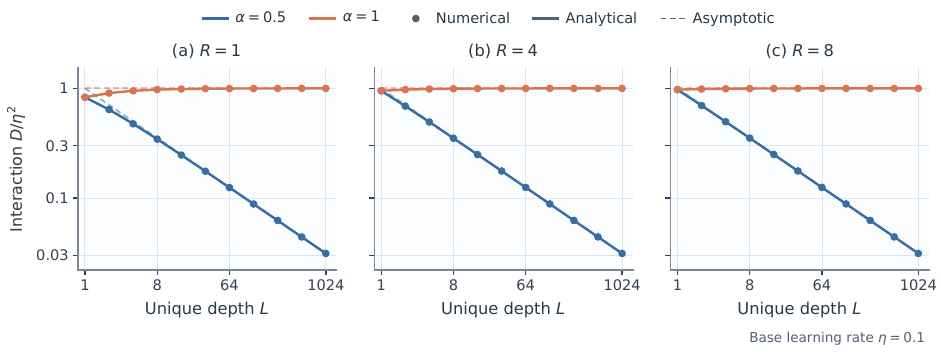}
  \caption{\textbf{Depth scaling and within-block parameter interactions.}
  Each point is a separate scalar model after one Adam update, with base LR
  $\eta=0.1$ and actual LR $\eta L^{\alpha-1}$.
  The normalized log-output discrepancy $D/\eta^2$
  (Eq.~\ref{eq:toy-depth-interaction}) isolates the effect of within-block
  parameter interactions while retaining the full $RL$-block composition.
  Numerical measurements (dots) agree with the exact finite-depth expression
  (solid lines); dashed lines show the asymptotic scaling $L^{\alpha-1}$.
  Both axes are logarithmic. The full 198-case sweep appears in
  Figure~\ref{fig:alpha-depth-toy-full}.}
  \label{fig:alpha-depth-toy}
\end{figure}

\para{Within-block interaction across depth.}
Our choice of $\alpha_L = 1$ is further motivated by within-block nonlinear parameter interactions across depth, following Depth-$\mu$P~\cite{yang2024tensor}.
Consider a scalar recurrence
\begin{equation}
    h_{r,\ell+1} = h_{r,\ell} + \frac{(a_\ell b_\ell - 1)h_{r,\ell}}{RL^\alpha} \,,
    \label{eq:scalar-recurrence}
\end{equation}
initialized at $a_\ell = b_\ell = 1$ after one Adam~\cite{kingma2014adam} update with learning rate $\eta L^{\alpha-1}$.
Comparing the full output $f_{\mathrm{full}}$ against a variant $f_{\mathrm{lin}}$ that removes only the within-block product $\Delta a_\ell \Delta b_\ell$, the log-output difference satisfies
\begin{equation}
  D = \log\frac{f_{\mathrm{full}}}{f_{\mathrm{lin}}},
  \qquad
  \frac{D}{\eta^2} \sim L^{\alpha-1} \quad (L \to \infty).
  \label{eq:toy-depth-interaction}
\end{equation}
As shown in Fig.~\ref{fig:alpha-depth-toy}, this within-block interaction term vanishes as $L$ grows for $\alpha = 0.5$ but remains $O(1)$ for $\alpha = 1$; full derivations and sweep results appear in Appendix~\ref{app:alpha-depth-toy}.

\para{\ours recipe.} 
We formally present~\ours in Table~\ref{tab:parameterization}, comparing against SP and Loop-$\mu$P~\cite{wang2026residual}.
We evaluate transfer by checking whether the optimal $\eta^{*}$ remains consistent across shapes alongside validation loss, and select task-specific settings (such as $\rho$, $\beta$) via small proxy scans across modalities.

\begingroup
\definecolor{ParamWidth}{HTML}{168343}
\definecolor{ParamDepth}{HTML}{1643C4}
\definecolor{ParamRecurrence}{HTML}{DB7100}
\newcommand{\widthfactor}[1]{\textcolor{ParamWidth}{#1}}
\newcommand{\depthfactor}[1]{\textcolor{ParamDepth}{#1}}
\newcommand{\recurrencefactor}[1]{\textcolor{ParamRecurrence}{#1}}
\begin{table}[!htbp]
  \centering
  \footnotesize
  \setlength{\tabcolsep}{3pt}
  \renewcommand{\arraystretch}{1.15}
  \caption{\textbf{\ours Parameterization.}
    We use $m_N=\frac{W}{W_0}$, $m_L=\frac{L}{L_0}$, and $m_R=\frac{R}{R_0}$.
    Green, blue, and orange identify width, unique-depth, and recurrence
    factors, respectively. Complete-P uses $\alpha_L=1$; the looped columns additionally show recurrence scaling.
    }
  \label{tab:parameterization}
  \begin{tabularx}{\textwidth}{@{}>{\raggedright\arraybackslash}p{0.195\textwidth}
    >{\hsize=0.82\hsize\raggedright\arraybackslash}X
    >{\hsize=0.94\hsize\raggedright\arraybackslash}X
    >{\hsize=1.06\hsize\raggedright\arraybackslash}X
    >{\hsize=1.18\hsize\raggedright\arraybackslash}X@{}}
    \toprule
    Parameterization & SP & Complete-P~\cite{dey2025completep} & Loop-$\mu$P~\cite{wang2026residual} & \ours \\
    \midrule
    Embedding init. var.
      & $\sigma_{\rm base}^{2}$
      & $\sigma_{\rm base}^{2}$
      & $\sigma_{\rm base}^{2}$
      & $\sigma_{\rm base}^{2}$ \\
    Embedding LR
      & $\eta_{\rm base}$
      & $\eta_{\rm base}$
      & $\eta_{\rm base}$
      & $\eta_{\rm base}$ \\
    \midrule
    Pre-norm init. var.
      & $\sigma_{\rm base}^{2}$
      & $\sigma_{\rm base}^{2}$
      & $\sigma_{\rm base}^{2}$
      & $\sigma_{\rm base}^{2}$ \\
    Pre-norm LR
      & $\eta_{\rm base}$
      & $\eta_{\rm base}$
      & $\eta_{\rm base}\depthfactor{m_L^{-\frac{1}{2}}}$
      & $\eta_{\rm base}$ \\
    \midrule
    Hidden weight init. var.
      & $\sigma_{\rm base}^{2}$
      & $\sigma_{\rm base}^{2}\widthfactor{m_N^{-1}}$
      & $\sigma_{\rm base}^{2}\widthfactor{m_N^{-1}}$
      & $\sigma_{\rm base}^{2}\widthfactor{m_N^{-1}}$ \\
    Hidden weight LR
      & $\eta_{\rm base}$
      & $\eta_{\rm base}\widthfactor{m_N^{-1}}$
      & $\eta_{\rm base}\widthfactor{m_N^{-1}}\depthfactor{m_L^{-\frac{1}{2}}}$
      & $\eta_{\rm base}\widthfactor{m_N^{-1}}$ \\
    Hidden bias LR
      & $\eta_{\rm base}$
      & $\eta_{\rm base}$
      & $\eta_{\rm base}\depthfactor{m_L^{-\frac{1}{2}}}$
      & $\eta_{\rm base}$ \\
    Hidden weight decay
      & $\lambda_{\rm base}$
      & $\lambda_{\rm base}\widthfactor{m_N}$
      & $\lambda_{\rm base}\widthfactor{m_N}$
      & $\lambda_{\rm base}\widthfactor{m_N}$ \\
    \midrule
    Attn residual ($z=$)
      & $\begin{aligned}[t]
          &x+\\[-2pt]
          &\mathrm{Attn}_{\ell}(\mathrm{Norm}(x))
        \end{aligned}$
      & $\begin{aligned}[t]
          &x+\depthfactor{m_L^{-1}}\\[-2pt]
          &\cdot\mathrm{Attn}_{\ell}(\mathrm{Norm}(x))
        \end{aligned}$
      & $\begin{aligned}[t]
          &x+\recurrencefactor{m_R^{-1}}\depthfactor{m_L^{-\frac{1}{2}}}\\[-2pt]
          &\cdot\mathrm{Attn}_{\ell}(\mathrm{Norm}(x))
        \end{aligned}$
      & $\begin{aligned}[t]
          &x+\recurrencefactor{m_R^{-1}}\depthfactor{m_L^{-1}}w_Av_r\\[-2pt]
          &\cdot\mathrm{Attn}_{\ell}(\mathrm{Norm}(x))
        \end{aligned}$ \\
    \addlinespace[3pt]
    FFN residual ($x^+=$)
      & $\begin{aligned}[t]
          &z+\\[-2pt]
          &\mathrm{FFN}_{\ell}(\mathrm{Norm}(z))
        \end{aligned}$
      & $\begin{aligned}[t]
          &z+\depthfactor{m_L^{-1}}\\[-2pt]
          &\cdot\mathrm{FFN}_{\ell}(\mathrm{Norm}(z))
        \end{aligned}$
      & $\begin{aligned}[t]
          &z+\recurrencefactor{m_R^{-1}}\depthfactor{m_L^{-\frac{1}{2}}}\\[-2pt]
          &\cdot\mathrm{FFN}_{\ell}(\mathrm{Norm}(z))
        \end{aligned}$
      & $\begin{aligned}[t]
          &z+\recurrencefactor{m_R^{-1}}\depthfactor{m_L^{-1}}w_Fv_r\\[-2pt]
          &\cdot\mathrm{FFN}_{\ell}(\mathrm{Norm}(z))
        \end{aligned}$ \\
    \midrule
    Final-norm init. var.
      & $\sigma_{\rm base}^{2}$
      & $\sigma_{\rm base}^{2}$
      & $\sigma_{\rm base}^{2}$
      & $\sigma_{\rm base}^{2}$ \\
    Final-norm LR
      & $\eta_{\rm base}$
      & $\eta_{\rm base}$
      & $\eta_{\rm base}$
      & $\eta_{\rm base}$ \\
    \midrule
    Unembedding init. var.
      & $\sigma_{\rm base}^{2}$
      & $\sigma_{\rm base}^{2}$
      & $\sigma_{\rm base}^{2}$
      & $\sigma_{\rm base}^{2}$ \\
    Unembedding LR
      & $\eta_{\rm base}$
      & $\eta_{\rm base}$
      & $\eta_{\rm base}$
      & $\eta_{\rm base}$ \\
    Unembedding forward
      & $x_{\rm out}W_{\rm unemb}^{\top}$
      & $x_{\rm out}W_{\rm unemb}^{\top}\widthfactor{m_N^{-1}}$
      & $x_{\rm out}W_{\rm unemb}^{\top}\widthfactor{m_N^{-1}}$
      & $x_{\rm out}W_{\rm unemb}^{\top}\widthfactor{m_N^{-1}}$ \\
    \midrule
    AdamW $\epsilon$ (blocks)
      & $\epsilon_{\rm base}$
      & $\epsilon_{\rm base}\widthfactor{m_N^{-1}}\depthfactor{m_L^{-1}}$
      & $\epsilon_{\rm base}\widthfactor{m_N^{-1}}\depthfactor{m_L^{-\frac{1}{2}}}$
      & $\epsilon_{\rm base}\widthfactor{m_N^{-1}}\depthfactor{m_L^{-1}}$ \\
    AdamW $\epsilon$ (emb./unemb.)
      & $\epsilon_{\rm base}$
      & $\epsilon_{\rm base}\widthfactor{m_N^{-1}}$
      & $\epsilon_{\rm base}\widthfactor{m_N^{-1}}$
      & $\epsilon_{\rm base}\widthfactor{m_N^{-1}}$ \\
    \bottomrule
  \end{tabularx}
  \par\smallskip
  \begin{minipage}{\textwidth}
    \footnotesize
    All LRs use AdamW. Complete-P uses the non-looped depth multiplier;
    for \ours, $L$ counts unique Core blocks.
    The branch coefficients in~\ours are $\gamma_{A,r}=g_{R,L}w_Av_r$ and
    $\gamma_{F,r}=g_{R,L}w_Fv_r$, with
    \[
      \begin{aligned}
        g_{R,L}&=m_R^{-\alpha_R}m_L^{-\alpha_L},
          &\alpha_R&=\alpha_L=1,\\
        w_A&=\rho\sqrt{\frac{2}{1+\rho^2}},
          &w_F&=\sqrt{\frac{2}{1+\rho^2}},\\
        v_r&=1+\beta\cos\!\left(\frac{\pi(r+\frac{1}{2})}{R}\right),
          &r&=0,\ldots,R-1.
      \end{aligned}
    \]
  \end{minipage}
\end{table}
\endgroup

\FloatBarrier

We first evaluate our~\ours HP transfer recipe on language modeling to showcase its effectiveness.
Then, we further apply~\ours to three vision tasks, 
including 3D reconstruction, class-to-image, and text-to-image,
across vision-to-vision and other-to-vision modalities.

\section{Experiments: Language modeling}
\label{sec:experiments_language}

We validate our proposed HP transfer recipe, \ours, on the language pretraining task. 
We first evaluate \ours on recurrence $R$ scaling,
and then we showcase the results in unique depth $L$ transfer.

\subsection{Setups}
\label{sec:language_pretrain}
\para{Models.}
Our language models use a Llama-style pre-norm decoder-only architecture~\cite{touvron2023llama} arranged in a $2/R\times L/2$ sandwich topology with width $768$ and $12$ attention heads.
By default, we use $L=4, R=1$ as a base model,
containing $105\text{M}$ unique parameters at $R=1$.
We set $\rho=2, \beta=0.5$ after the hyperparameter scan.
Further architectural details are given in Appendix~\ref{app:language-architecture}.
\para{Data.}
We split the Climb Mix~\cite{diao2025climb} dataset into a training set and a validation set.
We train our models on a fixed budget of $15.27\text{B}$ tokens.
We evaluate all models on the validation set with standard validation cross-entropy (CE) at the end of training.
\para{Training.}
During the training, we use a sequence length of $4{,}096$, and a batch size of $524{,}288$ tokens per update.
Models are trained with AdamW~\cite{loshchilov2019decoupled} ($(\beta_1,\beta_2)=(0.9,0.95)$, zero weight decay, and gradient norm clipping at $1.0$). All the training is paired with a $128$-step linear warmup and a final $10\%$ square-root decay to zero.
For each set of HPs, we use 5 different seeds for weight initialization.

\subsection{Transfer across recurrence $R$}
\label{sec:recurrence-lr-scaling}

\para{Full $R$-axis transfer.}
We further combine our HP transfer recipe with module-wise residuals (Sec.~\ref{sec:module-residuals}) and visit-wise residuals (Sec.~\ref{sec:visit-residuals}).
Figure~\ref{fig:teaser-hp-transfer} shows the full method across the $R$-axis.
With all components active, our recipe maintains stable HP transfer across recurrence counts and achieves lower validation CE than Loop-$\mu$P~\cite{wang2026residual}.
The isolated LR-scaling analysis and supporting fits are presented in Appendix~\ref{app:recurrence-lr-analysis}.

\subsection{Transfer across unique depth $L$ and mixed $R,L$-axes.}
We next ask whether the reference model's LR remains effective as unique Core depth $L$ grows, first at fixed recurrence and then jointly with $R$.
We compare \ours with our sandwich adaptation of Loop-$\mu$P~\cite{wang2026residual}, sweeping each recipe in its own base LR $\eta_{\rm base}$ in Table~\ref{tab:parameterization}.

\begingroup
\setlength{\intextsep}{5pt}
\begin{figure}[!htbp]
  \centering
  \captionsetup{skip=4pt}
  \captionsetup[subfigure]{skip=2pt}
  \begin{subfigure}[t]{0.36\textwidth}
    \centering
    \includegraphics[width=\linewidth]{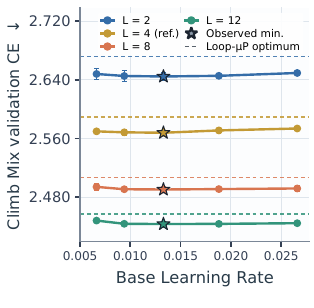}
    \caption{\ours (ours).}
    \label{fig:depth-transfer-b-axis}
  \end{subfigure}\hspace{0.04\textwidth}%
  \begin{subfigure}[t]{0.36\textwidth}
    \centering
    \includegraphics[width=\linewidth]{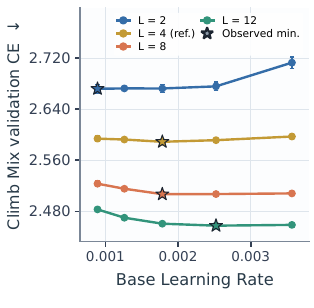}
    \caption{Loop-$\mu$P.}
    \label{fig:depth-transfer-loopmup-b-axis}
  \end{subfigure}
  \caption{\textbf{Transferring across unique depth $L$.}
    At fixed recurrence, \ours retains a common best tested LR as the Core deepens, whereas Loop-$\mu$P~\cite{wang2026residual} shows less consistent alignment.
    Dashed lines in (a) mark each matching model's lowest mean Climb Mix~\cite{diao2025climb} validation CE over the Loop-$\mu$P sweep in (b).}
  \label{fig:depth-transfer}
\end{figure}
\endgroup

\para{Transfer across unique depth $L$-axis.}
At fixed recurrence, the \ours minima in Figure~\ref{fig:depth-transfer-b-axis} stay at the reference LR as deeper Cores reach lower CE.
The Loop-$\mu$P minima in Figure~\ref{fig:depth-transfer-loopmup-b-axis} are less consistently aligned.
To assess whether the advantage survives retuning, the dashed references in Figure~\ref{fig:depth-transfer-b-axis} show each matching Loop-$\mu$P model's best measured mean CE over its LR sweep.
The \ours curves stay below these references throughout the grid, combining LR reuse with lower loss even against a separately tuned baseline.
\begingroup
\setlength{\intextsep}{5pt}
\begin{figure}[!htbp]
  \centering
  \captionsetup{skip=4pt}
  \captionsetup[subfigure]{skip=2pt}
  \begin{subfigure}[t]{0.36\textwidth}
    \centering
    \includegraphics[width=\linewidth]{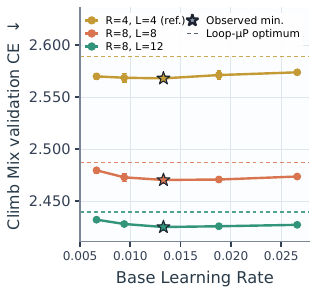}
    \caption{\ours (ours).}
    \label{fig:depth-transfer-c-axis}
  \end{subfigure}\hspace{0.04\textwidth}%
  \begin{subfigure}[t]{0.36\textwidth}
    \centering
    \includegraphics[width=\linewidth]{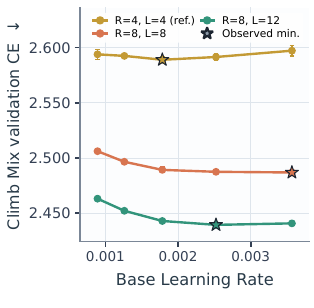}
    \caption{Loop-$\mu$P.}
    \label{fig:depth-transfer-loopmup-c-axis}
  \end{subfigure}
  \caption{\textbf{Transferring jointly across $(R,L)$.}
    Increasing both recurrence and unique depth preserves the best tested LR for \ours, while the Loop-$\mu$P~\cite{wang2026residual} target minima shift toward larger LRs.
    Dashed lines in (a) show the matching models' lowest mean Climb Mix~\cite{diao2025climb} validation CE in (b), so the lower \ours curves reflect an advantage even after retuning the baseline.}
  \label{fig:mixed-transfer}
\end{figure}
\endgroup

\para{Transferring across both $R,L$ axes.}
Successful transfer along either axis alone does not guarantee that the same LR remains effective when both change together.
Figure~\ref{fig:mixed-transfer} tests this joint change by increasing both unique Core depth and recurrence from the same reference model.
In Figure~\ref{fig:depth-transfer-c-axis}, the \ours target curves move downward while retaining their lowest measured means at the reference LR.
The benefit of the larger models is therefore accessible without a new LR choice.
By contrast, the Loop-$\mu$P target minima in Figure~\ref{fig:depth-transfer-loopmup-c-axis} move to the right of the reference minimum, favoring larger base LRs within that recipe.
One target reaches its lowest measured loss at the edge of the grid, so the sweep does not locate its optimum beyond that boundary.
Nevertheless, even the best tested Loop-$\mu$P losses, shown as horizontal references in Figure~\ref{fig:depth-transfer-c-axis}, remain above the matching \ours curves.
The combined change in $(R,L)$ therefore preserves both the shared minimizing LR and the validation-loss advantage of \ours over the displayed shapes.

\section{Experiments: 3D reconstruction}
\label{sec:reconstruction-experiments}

We next ask whether recurrent computation can preserve the visual reasoning needed for 3D reconstruction while reducing the number of learned parameters.
As a vision-to-vision task, it takes references images as the input, and outputs estimated camera poses and novel-view synthesis (NVS).

\subsection{Setups}
\para{Models.}

We apply the sandwich design to both the pose and geometry encoders of E-RayZer~\cite{zhao2026erayzer}, retaining its prediction heads and renderer.
Each looped encoder uses $P=C=4$ and $L=2$.
At $R=4$, its effective depth is 16 layers, matching its non-looped counterpart; increasing $R$ to eight raises this depth to 24 layers without adding parameters.
We compare naive looping with our residual allocation at $R=4$, then examine the latter at $R=8$.

\para{Data.}
We use the DL3DV~\cite{ling2024dl3dv} training split, and hold out the 140-scene benchmark for evaluation.
Each scene window contains ten $256\times256$ views, split into five reference and five target views.

\para{Training.}
We keep E-RayZer's reconstruction objective and AdamW optimizer, training for 152K iterations with a global batch of 192 scene windows.
All models use the same peak LR and warmup--cosine schedule, allowing the comparison to focus on sharing and recurrent computation under the native training recipe.
Task-specific residual scales and further training details are given in Appendix~\ref{app:vision-protocols}.

\para{Evaluation.}
We measure relative-pose accuracy (RPA)~\cite{zhao2026erayzer} and novel-view PSNR.
Compute counts encoder forward operations on the same scene window, excluding rendering and backward computation.

\subsection{Reconstruction with shared parameters}
\begin{table}[!htbp]
  \centering
  \small
  \setlength{\tabcolsep}{3pt}
  \renewcommand{\arraystretch}{1.12}
  \caption{\textbf{Parameter sharing for 3D reconstruction.}
    E-RayZer~\cite{zhao2026erayzer} on the DL3DV~\cite{ling2024dl3dv} test split.
    At $R=4$, looping improves the reported pose and rendering metrics with $34.6\%$ fewer parameters at matched encoder compute; our residual allocation gives a further improvement over the naive loop.
    $R=8$ achieves the highest displayed scores with the same parameters and $1.5\times$ encoder compute.
    $R$ counts Core visits.
    TFLOPs count encoder forward matrix operations per 10-view $256^2$ scene window. Rel. normalizes them to the non-looped model.
    \textbf{Bold} and \underline{underlining} mark the best and second-best displayed values per metric, respectively.
  }
  \label{tab:erayzer-loop-comparison}
  \begin{tabular*}{\linewidth}{@{\extracolsep{\fill}}lcrrrrrr@{}}
    \toprule
    Method & $R$ & Model size & TFLOPs & Rel.
      & RPA@5 (\%) $\uparrow$
      & RPA@15 (\%) $\uparrow$
      & PSNR (dB) $\uparrow$ \\
    \midrule
    Non-looped
      & 1 & 245.83M & 1.114 & $1.000\times$ & 74.60 & 89.90 & 20.407 \\
    Looped (naive loop)
      & 4 & 160.88M & 1.114 & $1.000\times$ & 77.46 & { 90.48} & {20.519} \\
    Looped (Ours)
      & 4 & 160.88M & 1.114 & $1.000\times$ & \underline{77.70} & \underline{90.52} & \underline{20.525} \\
    Looped (Ours)
      & 8 & 160.88M & 1.670 & $1.500\times$ & \textbf{79.16}  & \textbf{90.79} & \textbf{20.684} \\
    \bottomrule
  \end{tabular*}
\end{table}

\begin{wrapfigure}{R}{0.49\textwidth}
  \vspace{-0.5\intextsep}
  \centering
  \includegraphics[width=\linewidth]{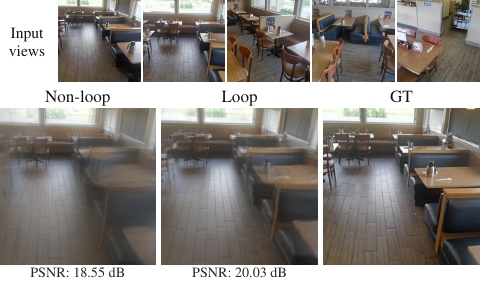}
  \caption{\textbf{Qualitative comparison between non-loop and loop with E-RayZer~\cite{zhao2026erayzer} on the DL3DV~\cite{ling2024dl3dv} test split.}
  NVS from the looped model fixes the structural errors on the floor.}
  \label{fig:erayzer-paired-comparison}
  \vspace{-0.5\baselineskip}
\end{wrapfigure}

\para{Sharing at matched encoder compute.}
Table~\ref{tab:erayzer-loop-comparison} shows that recurrence can reduce model size without sacrificing either pose estimation or rendering quality.
At $R=4$, our model uses $34.6\%$ fewer parameters at the same encoder compute as the non-looped model, while improving both pose accuracy and PSNR.
The naive loop already captures much of this gain; our residual allocation yields a smaller further improvement across all three reported metrics.
The evidence therefore supports parameter sharing as an effective design for this reconstruction task, with an additional benefit from how residual computation is allocated.
Figure~\ref{fig:teaser-right} complements the quantitative comparison with matched target views: the looped renderings preserve the overall scene structure, including the building facade and foreground objects, while both models still lose fine detail relative to ground truth.
Figure~\ref{fig:erayzer-paired-comparison} shows the conditioning images and per-view PSNR for one matched non-loop/loop comparison.

\para{More computation with the same parameters.}
Increasing recurrence to $R=8$ improves both pose accuracy and PSNR again, without enlarging the model.
This separates the role of parameter count from the amount of computation: the shared Core can make productive use of further visits.
The improvement comes with $1.5\times$ encoder compute, however, so it describes a second operating point rather than a gain at the original budget.
\FloatBarrier

\section{Experiments: Class-to-Image generation}
\label{sec:class-image-experiments}

Image generation asks a different question of shared computation: can a smaller denoiser retain the quality and diversity of the generated distribution?
We use this setting to distinguish the benefits of parameter sharing from those of spending more computation on the same shared weights.

\subsection{Setups}
\para{Models.}
We use PixelDiT-XL~\cite{yu2026pixeldit}, applying recurrence to its patch-level Transformer while retaining the four pixel-level blocks.
The patch-level Transformer uses $P=8$, $L=4$, and $C=2$.
At $R=4$, its effective depth is 26 layers, matching the non-looped model with fewer unique weights; increasing $R$ to eight raises the effective depth to 42 layers.
This provides a compute-matched comparison and a higher-compute comparison at the same looped model size.

\para{Data.}
Following PixelDiT~\cite{yu2026pixeldit}, we train on the full ImageNet-1K training set~\cite{deng2009imagenet}, containing approximately 1.28M images from 1,000 classes, at $256\times256$ resolution.
Class labels provide the generation condition.

\para{Training.}
We follow PixelDiT's flow-matching objective with representation alignment (REPA)~\cite{yu2025repa} and use AdamW with a global batch of 256 images.
The quantitative endpoints in Table~\ref{tab:benchmark-400320} are evaluated after 400,320 updates, so additional recurrence increases computation per image rather than the number of training images seen.

\para{Evaluation.}
Following PixelDiT~\cite{yu2026pixeldit}, we evaluate 50K samples from the EMA model using the ADM evaluation suite~\cite{dhariwal2021diffusion}.
We report Fr\'echet Inception Distance (gFID)~\cite{heusel2017gans}, spatial FID (sFID)~\cite{nash2021sparse}, Inception Score (IS)~\cite{salimans2016improved}, and improved precision and recall~\cite{kynkaanniemi2019improved}.
These measures provide complementary views of fidelity and coverage, so we examine their agreement rather than relying on a single score.

\subsection{Results}
\begin{table}[!htbp]
  \centering
  \small
  \setlength{\tabcolsep}{3pt}
  \renewcommand{\arraystretch}{1.12}
  \caption{\textbf{Parameter savings and generation quality with PixelDiT.}
    Non-looped and looped PixelDiT-XL~\cite{yu2026pixeldit} on ImageNet-256~\cite{deng2009imagenet}.
    The $R=4$ loop uses $36.0\%$ fewer parameters at matched nominal training compute, with higher gFID/sFID and similar recall.
    The $R=8$ endpoint improves gFID/sFID and recall over the non-looped model at $1.413\times$ compute.
    $R$ counts Core visits.
    Rel. is normalized to the non-looped model.
    We report gFID~\cite{heusel2017gans}, sFID~\cite{nash2021sparse}, IS~\cite{salimans2016improved}, and precision/recall~\cite{kynkaanniemi2019improved} using the ADM evaluation protocol~\cite{dhariwal2021diffusion}.
    \textbf{Bold} and \underline{underlining} mark the best and second-best displayed values per metric, respectively.
  }
  \label{tab:benchmark-400320}
  \begin{tabular*}{\linewidth}{@{\extracolsep{\fill}}lcrrrrrrrr@{}}
    \toprule
    Method & $R$ & Model size & TFLOPs & Rel.
      & gFID $\downarrow$
      & sFID $\downarrow$
      & IS $\uparrow$
      & Precision $\uparrow$
      & Recall $\uparrow$ \\
    \midrule
    Non-looped
      & 1 & 797.38M & 0.989 & $1.000\times$ & \underline{2.2583} & \underline{4.9459}
      & \textbf{64.4651} & \textbf{0.7898} & 0.5990 \\
    Naive Loop
      & 4 & 510.60M & 0.989 & $1.000\times$ & 2.8348 & 5.1585
      & 59.3902 & 0.7745 & 0.5904 \\
    Loop
      & 4 & 510.60M & 0.989 & $1.000\times$ & 2.7511 & 5.2541
      & 59.8168 & \underline{0.7776} & \underline{0.5999} \\
    Loop
      & 8 & 510.60M & 1.398 & $1.413\times$ & \textbf{2.2423} & \textbf{4.9312}
      & \underline{63.6385} & 0.7713 & \textbf{0.6057} \\
    \bottomrule
  \end{tabular*}
\end{table}

\para{Compute-matched comparison.}
Table~\ref{tab:benchmark-400320} shows that the $R=4$ loop reduces model size by $36.0\%$, and it achieves comparable performance as its non-looped model under the same computes.
The paired samples in Figure~\ref{fig:teaser-middle} show recognizable class-conditioned content from both models; the distribution-level metrics reveal differences that a few selected images cannot resolve.

\begin{figure}[!htbp]
  \centering
  \includegraphics[width=\linewidth]{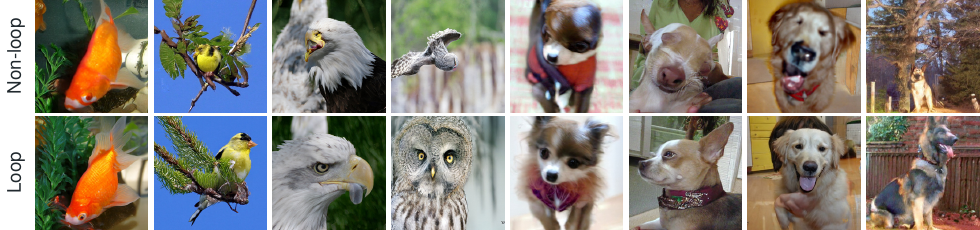}
  \caption{\textbf{Qualitative comparison on PixelDiT~\cite{yu2026pixeldit}}.
  We use $R=4$ for the looped model. Two model share the same FLOPs.
  We generate paired images with same initial noise. 
  The looped model fixes structural errors present in the non-looped model.
  }
  \label{fig:pixeldit-paired-comparison}
\end{figure}

\para{Recovering quality through additional recurrence.}
With $R=8$, the looped model outperforms the non-looped model on gFID, sFID, and recall. It also achieves comparable IS and precision.
This suggests that with more recurrences, the quality of looped model can be further improved.

\para{Paired qualitative comparison.}
We show our looped model ($R=4$) with its non-looped counterpart in Figure~\ref{fig:pixeldit-paired-comparison}. Two models share the same inference FLOPs, matched class labels, random seeds, and initial noise.
\FloatBarrier

\section{Experiments: Text-to-Image generation}
\label{sec:text-image-experiments}

We finally examine whether a compact looped denoiser can retain compositional text-to-image generation ability through both pretraining and supervised fine-tuning (SFT).
Comparing these two stages lets us ask whether an advantage in the pretrained model persists after adaptation.

\subsection{Setups}
\para{Models.}
We build on MiniT2I-B/32~\cite{minit2i2026}, a pixel-space denoiser conditioned on frozen FLAN-T5-Large~\cite{chung2024scaling} text features, generating $512\times512$ images.
The looped models use $P=C=6$, $L=5$, and $R=4$, giving 17 unique blocks and an effective depth of 32 layers.
We compare naive and \ours loops with non-looped models of 17 and 32 blocks: the former matches their $260\text{M}$ parameter count, while the latter matches their nominal denoiser compute with $473\text{M}$ parameters.
This separates comparison at a common model size from comparison at a common arithmetic budget.

\para{Data.}
Pretraining uses approximately 10.95M downloaded CC12M~\cite{changpinyo2021cc12m} images with LLaVA-NeXT recaptions~\cite{captionemporium2024cc12m}, reserving disjoint subsets for validation and held-out reporting.
For SFT, we follow MiniT2I's mixture of BLIP3o-60K~\cite{chen2025blip3o}, DALL-E 3-generated image--text pairs~\cite{betker2023dalle3} from OpenDatasets~\cite{opendatasets2024dalle3}, and ShareGPT-4o-Image~\cite{chen2025sharegpt4o}, with 119,239 available examples.

\para{Training.}
Each model is pretrained for 250K updates and then fine-tuned for another 40K updates, using AdamW~\cite{loshchilov2019decoupled} and a global batch of 1,024 throughout.
SFT continues the pretrained model and optimizer states on the new data mixture, so the before/after comparison measures adaptation of the same training run.
The objective, task-specific LR settings, and sampling details are given in Appendix~\ref{app:vision-protocols}.

\para{Evaluation.}
We report GenEval~\cite{ghosh2023geneval} and DPG-Bench~\cite{hu2024ella} before and after SFT to assess compositional prompt following, and post-SFT FID~\cite{heusel2017gans} to assess image-distribution quality.
FID is measured after SFT on 30K COCO~\cite{lin2014coco} samples for EMA weights.

\subsection{Results}
\begin{table}[!htbp]
  \centering
  \small
  \setlength{\tabcolsep}{3pt}
  \renewcommand{\arraystretch}{1.12}
  \caption{\textbf{Compositional text-to-image generation before and after SFT.}
    MiniT2I-B/32~\cite{minit2i2026} at $512^2$ resolution. 
    arrows ($\rightarrow$) connect 250K CC12M~\cite{changpinyo2021cc12m} pretraining updates to an additional 40K updates on a mixture of BLIP3o-60K~\cite{chen2025blip3o}, DALL-E 3~\cite{opendatasets2024dalle3}, and ShareGPT-4o-Image~\cite{chen2025sharegpt4o}.
    At the same parameter count and nominal compute, \ours exceeds naive looping on GenEval~\cite{ghosh2023geneval} and DPG-Bench~\cite{hu2024ella} (DPG) at both stages.
    After SFT, it also exceeds the larger dense model on these benchmarks with $45.0\%$ fewer denoiser parameters, although the dense model has a slightly lower FID~\cite{heusel2017gans}.
    FD-DINOv2~\cite{stein2023exposing} and CMMD~\cite{jayasumana2024rethinkingfid} compare 30K generated images from the SFT EMA models with the corresponding COCO reference set.
    $R$ counts Core visits ($R=1$ for non-looped models). 
    \textbf{Bold} and \underline{underlining} mark the best and second-best results per metric, respectively; pretraining and SFT are ranked separately.
  }
  \label{tab:pretrain-sft}
  \begin{tabular*}{\linewidth}{@{\extracolsep{\fill}}lcrrrrrrrr@{}}
    \toprule
    Method & $R$ & Model size & TFLOPs & Rel. & GenEval ($\%$) $\uparrow$ & DPG $\uparrow$ & FID $\downarrow$ & FD-DINOv2 $\downarrow$ & CMMD $\downarrow$ \\
    & & & & & Pretrain $\to$ SFT & Pretrain $\to$ SFT & SFT & SFT & SFT \\
    \midrule
    Non-looped & 1 & 260M & 0.438 & $0.543\times$ & $53.34 \to 84.46$ & $78.07 \to 82.63$ & $ 27.63$ & $281.740$ & $1.0301$ \\
    Naive loop & 4 & 260M & 0.806 & $1.000\times$ & $51.39 \to 82.38$ & $78.56 \to 83.33$ & $\underline{27.28}$ & $296.525$ & $1.0223$ \\
    Loo-P (ours) & 4 & 260M & 0.806 & $1.000\times$ & $\mathbf{55.21} \to \mathbf{85.83}$ & $\mathbf{80.63} \to \mathbf{84.67}$ & $ 27.62$ & $\mathbf{274.819}$ & $\mathbf{0.9596}$ \\
    Non-looped & 1 & 473M & 0.806 & $1.000\times$ & $\underline{53.58} \to \underline{85.29}$ & $\underline{80.53} \to \underline{83.86}$ & $\mathbf{26.32}$ & $\underline{278.751}$ & $\underline{0.9722}$ \\
    \bottomrule
  \end{tabular*}

\end{table}

\para{Beyond naive parameter sharing.}
In Table~\ref{tab:pretrain-sft}, \ours exceeds the naive loop on both GenEval and DPG before SFT, and this ordering persists after finetuning.
The post-SFT gains are $3.45 \%$ on GenEval and $1.34 \%$ on DPG.
Within this comparison, the \ours recipe makes more effective use of the shared model than naive looping.

\begin{figure}[!htbp]
  \centering
  \includegraphics[width=\linewidth]{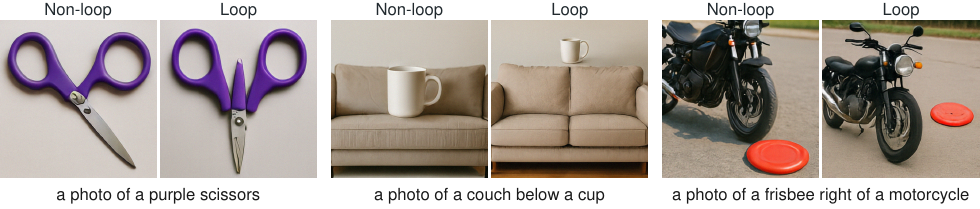}
  \caption{\textbf{Qualitative comparison between non-looped and looped models on MiniT2I~\cite{minit2i2026}.}
  We compare our looped model ($R=4$) against a compute-matched non-looped baseline using the same prompts and initial noise. The non-looped baseline struggles with spatial relationships and exhibits structural artifacts (e.g., the scissors).
  }
  \label{fig:minit2i-paired-comparison}
\end{figure}

\para{Comparison at matched compute.}
The advantage also extends to the larger non-looped reference: whether before or after SFT, \ours scores higher on both GenEval and DPG despite with substantially fewer denoiser parameters than the non-looped model.
This further demonstrates our~\ours works effectively across different tasks.
Figure~\ref{fig:minit2i-paired-comparison} presents three selected non-loop/loop pairs with their corresponding prompts.

\FloatBarrier

\section{Discussion: loops for different modalities}
\label{sec:cross-modality-analysis}

\begin{figure}[!htbp]
  \centering
  {\small
    \textcolor[HTML]{0072B2}{\rule[0.5ex]{1.5em}{1pt}}\;Non-looped\qquad
    \textcolor[HTML]{777D87}{\rule[0.5ex]{1.5em}{1pt}}\;Naive loop\qquad
    \textcolor[HTML]{D55E00}{\rule[0.5ex]{1.5em}{1pt}}\;Loo-P
    \par}
  \vspace{2pt}
  \captionsetup[subfigure]{font=footnotesize,skip=2pt}
  \begin{subfigure}[t]{0.243\linewidth}
    \includegraphics[width=\linewidth]{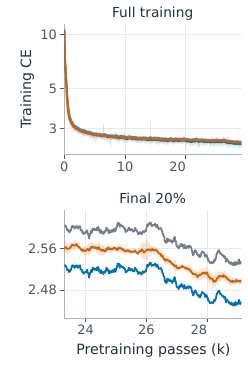}
    \caption{Language}
    \label{fig:training-loss-language}
  \end{subfigure}\hfill
  \begin{subfigure}[t]{0.243\linewidth}
    \includegraphics[width=\linewidth]{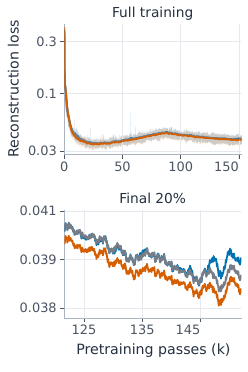}
    \caption{E-RayZer}
    \label{fig:training-loss-erayzer}
  \end{subfigure}\hfill
  \begin{subfigure}[t]{0.243\linewidth}
    \includegraphics[width=\linewidth]{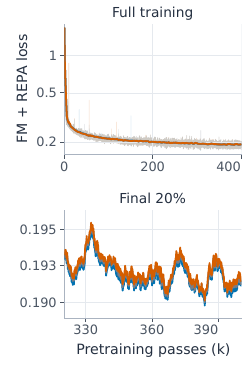}
    \caption{PixelDiT}
    \label{fig:training-loss-pixeldit}
  \end{subfigure}\hfill
  \begin{subfigure}[t]{0.243\linewidth}
    \includegraphics[width=\linewidth]{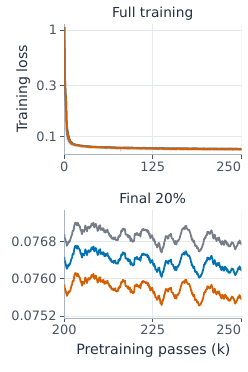}
    \caption{MiniT2I}
    \label{fig:training-loss-minit2i}
  \end{subfigure}
  \caption{\textbf{The training-loss tradeoff of parameter sharing depends on the task.}
    We compare compute-matched Non-looped, Naive loop, and \ours on (a)~language, (b)~E-RayZer~\cite{zhao2026erayzer}, (c)~PixelDiT~\cite{yu2026pixeldit}, and (d)~MiniT2I~\cite{minit2i2026}.
    \ours acheives comparable or lower loss in vision tasks (b),(c),(d).
    \ours reduces but does not close the gap on language CE.
    }
  \label{fig:training-loss-modalities}
\end{figure}

One of the benefits of Looped Transformers is that we can replace unique layers with repeated layers under a matched compute budget while using fewer model parameters.
For language modeling, however, outperforming non-looped baselines at matched compute remains challenging~\cite{schwethelm2026recurrence,saunshi2025latent}, with reported gains depending on the task and carefully designed architectures or training recipes~\cite{alabdulmohsin2025rins,wang2026smelt}.
Our language experiments also reflect this challenge in Table~\ref{tab:language-dense-loop} and Figure~\ref{fig:training-loss-language}.
Although \ours reduces the gap between naive looping and the non-looped baseline, it still shows a higher CE loss.
This raises a broader question: does a similar gap exist across other modalities, \textit{e.g.}, vision?

To answer this question, we study the pretraining losses for different vision tasks and report them in Figure~\ref{fig:training-loss-modalities}. 

\Needspace{6\baselineskip}
\subsection{Vision: from reconstruction to generation}
\label{sec:modality-vision}
We have studied multi-view 3D reconstruction, text-to-image (T2I) generation, and class-to-image (C2I) generation under \textbf{a FLOPs-matched setting}. 
Our key observation is
\begin{itemize}
    \item \textbf{3D Reconstruction (E-RayZer):} Naive looping has already \textbf{outperformed} the non-looped model. \ours further boosts the performance (Table~\ref{tab:erayzer-loop-comparison}, Figure~\ref{fig:training-loss-erayzer}).
    \item \textbf{T2I (MiniT2I):} \ours improves overall performance and it outperforms the non-looped counterpart. (Table~\ref{tab:pretrain-sft}). \ours also shows a lower pretraining loss (Figure~\ref{fig:training-loss-minit2i}).
    \item \textbf{C2I (PixelDiT):} Although \ours improves overall metrics, it is slightly worse than the non-looped model (Table~\ref{tab:benchmark-400320}). For pretraining, \ours shows a comparable training loss with the non-looped model (Figure~\ref{fig:training-loss-pixeldit}). 
\end{itemize}
\para{Iterative refinement and input informativeness.}
Across vision tasks, looping is more naturally aligned with settings where the input already provides rich structural or compositional information to iteratively refine, rather than expanding sparse conditions or relying heavily on distinct layer capacity.
In 3D reconstruction (E-RayZer), multi-view observations supply dense geometric constraints across views, helping a shared Core improve reconstruction loss and geometry metrics with fewer parameters.
T2I (MiniT2I) receives richer compositional prompts alongside a frozen text encoder, 
whereas C2I (PixelDiT) starts from a sparse class label. Consequently, shared blocks can preserve or improve semantic and alignment signals while fine-grained generation quality and distributional fidelity remain more sensitive to independent layer capacity.

\begin{table}[!htbp]
  \centering
  \begin{minipage}{0.88\linewidth}
  \small
  \setlength{\tabcolsep}{4pt}
  \renewcommand{\arraystretch}{1.12}
  \caption{\textbf{Parameter savings with a remaining language-modeling loss gap.}
    Climb Mix~\cite{diao2025climb} validation CE after $15.27\text{B}$ training tokens (lower is better).
    All models have an effective depth of 20 layers at width 768 and use the same tokenizer and validation inputs.
  }
  \label{tab:language-dense-loop}
  \begin{tabular*}{\linewidth}{@{\extracolsep{\fill}}lrrr@{}}
    \toprule
    Model & Parameters & Effective depth & Val. CE $\downarrow$ \\
    \midrule
    Non-looped & 190.74M & 20 & \textbf{2.53166} \\
    Naive loop ($R=4,L=4$) & 105.79M & 20 & 2.61374 \\
    \ours ($R=4,L=4$) & 105.79M & 20 & 2.56796 \\
    \bottomrule
  \end{tabular*}
  \end{minipage}
\end{table}

\subsection{Language retains a gap to independent layers.}
\label{sec:modality-language}
Language presents a less favorable case under standard pretraining objectives.
Although \ours{} improves on naive looping, its final-window training CE remains $1.53\%$ above the non-looped baseline in Figure~\ref{fig:training-loss-language}, and Table~\ref{tab:language-dense-loop} shows that while it removes more than half of the naive loop's validation CE gap, it does not close it despite similar parameter savings to MiniT2I.

\begin{figure}[!ht]
  \centering
  \includegraphics[width=\linewidth,trim={0 27bp 0 51bp},clip]{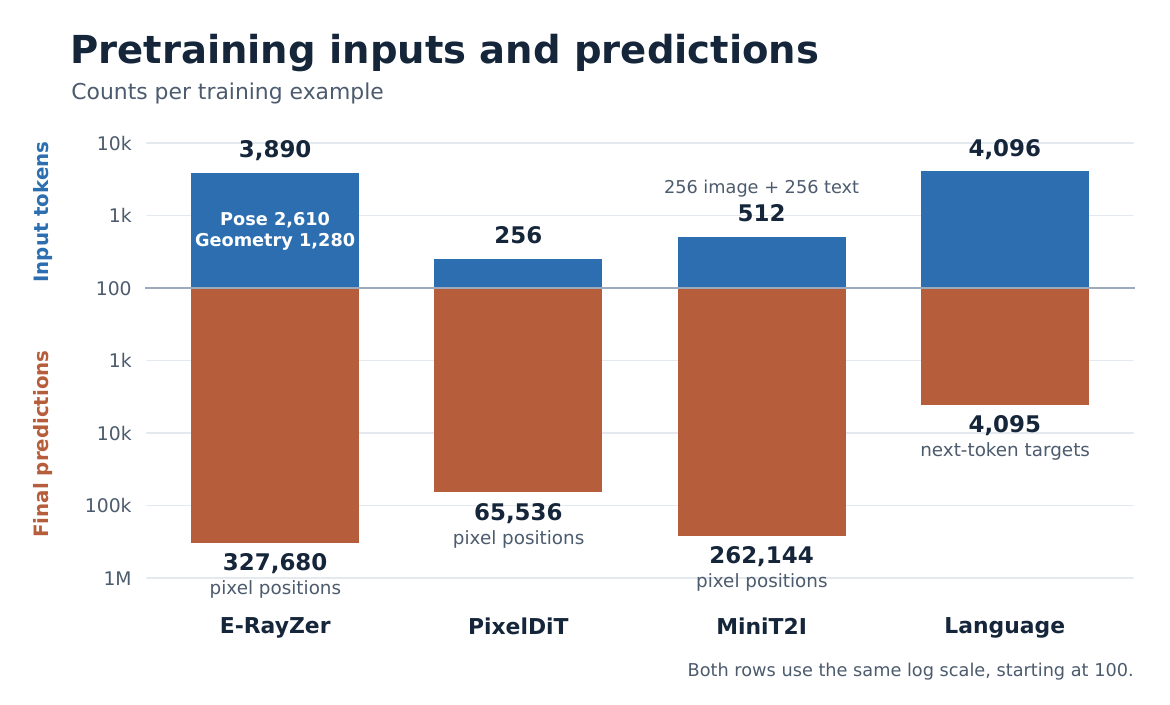}
  \caption{\textbf{Inputs and predictions across pretraining tasks.}
    Both rows use the same logarithmic scale with a baseline of 100.
    Vision pretraining takes less tokens as input, but commonly predicts far more rendered pixels for loss.
    }
  \label{fig:pretraining-io-counts}
\end{figure}

This contrast aligns with differences in sequence structure and prediction heads across tasks, as illustrated by the input and output counts in Figure~\ref{fig:pretraining-io-counts}.
In our vision setups, 
the model usually outputs dense pixels,
\textit{e.g.}, E-Rayzer predicts $327{,}680$ pixels while it only takes $3{,}890$ tokens.
By contrast, our language setup processes $4{,}096$-token sequences under causal masking and predicts $4{,}095$ next-token targets directly over a $32{,}000$-token vocabulary.
Because every position must simultaneously support multi-scale contextual routing and fine-grained discrete classification without a spatial decoding stage, standard next-token pretraining places heavier demands on independent layer capacity.
Looping may become comparatively more advantageous in language when tasks shift toward multi-step reasoning, post-training, or test-time compute scaling where latent state iteration can be leveraged more directly, which remains an open hypothesis beyond our pretraining results.

Taken together, these results show a more favorable training-loss tradeoff for sharing in the tested vision tasks than in language pretraining.

\FloatBarrier
\section{Related Work}
\label{sec:related}
\para{Looped Transformers.}
The development of Looped Transformers connects two goals: sharing parameters across depth and allocating additional computation to an input.
Adaptive Computation Time~\cite{graves2016adaptive} introduced learned halting for recurrent networks; Universal Transformers~\cite{dehghani2018universal} subsequently combined shared self-attentive transformations with position-dependent halting, while ALBERT~\cite{lan2019albert} used cross-layer sharing to reduce language-model size.
Work in 2023--2024 connected looping to programmable computation~\cite{giannou2023programmable} and learning iterative in-context algorithms~\cite{yang2024learningalgorithms}, highlighting an inductive bias beyond parameter compression.
Relaxed Recursive Transformers~\cite{bae2024relaxed} then explored converting pretrained language models into shared architectures with layer-specific LoRA adaptations.
In 2025, Huginn~\cite{geiping2025huginn} scaled recurrent-depth language pretraining with a Prelude--Core--Coda architecture and variable test-time computation, Mixture-of-Recursions~\cite{bae2025mor} introduced token-level recursion routing, and Ouro~\cite{zhu2025ouro} combined large-scale looped pretraining with learned depth allocation.
These works shifted attention toward how shared computation should be organized and trained.
In 2026, Hyperloop~\cite{zeitoun2026hyperloop} enriched loop boundaries with multiple residual streams, LT2~\cite{deng2026lt2} paired recurrence with efficient attention, and Loopie~\cite{gao2026loopie} studied consecutive repetition of individual layers in MoE models.
A recent technical report~\cite{huang2026loopedmodels} further separates the effects of recurrence placement, input injection, and recurrent-state design, providing evidence for sandwich topologies.
Against this background, we study how residual allocation and hyperparameter scaling jointly improve training within a shared Core.

\para{Looped Transformers for vision.}
Visual applications extend this trajectory from recognition and abstract reasoning to generation.
RINS~\cite{alabdulmohsin2025rins} demonstrated recursive inference scaling in language and multimodal models, including image--text recognition.
HRM~\cite{wang2025hrm} and its simpler successor TRM~\cite{jolicoeur2025trm} studied recurrent computation on puzzles and ARC, while LoopViT~\cite{shu2026loopvit} applied a shared convolution--attention block with adaptive halting to visual ARC.
These studies motivate iterative visual reasoning, but their discrete puzzle tasks differ from reconstruction and image synthesis.
For visual generation, ELT~\cite{goyal2026elt} uses intra-loop self-distillation to train models that support multiple inference depths.
Looped-DiT~\cite{chng2026loopeddit} combines intermediate-loop supervision with self-modulating attention, and LiFT~\cite{derakhshani2026lift} supervises a continuous depth trajectory to support longer recurrent rollouts.
Dooop~\cite{wang2026dooop} likewise supports adjustable inference depth, but obtains its looped model by distilling a pretrained non-looped diffusion Transformer.
Whereas LiFT anchors intermediate targets to the initial prediction and the flow-matching target, Dooop constructs each target from the preceding loop's prediction toward the teacher output.
Here, repeating shared layers within a denoiser evaluation provides a computation axis distinct from increasing the number of denoising steps.
Our study connects residual design and HP transfer with evaluations on 3D reconstruction and image generation, and compares their parameter-sharing trade-offs with language pretraining under task-specific compute matching.

\para{Hyperparameter transfer.}
The $\mu$P framework~\cite{yang2022transfer} established that suitable parameterization can preserve effective hyperparameters as width grows, allowing small models to guide larger ones.
Depth-$\mu$P~\cite{yang2024tensor} extended this question to residual depth through coordinated scaling of residual branches and parameter updates.
CompleteP~\cite{dey2025completep} further showed why HP transfer alone is insufficient: a parameterization should also preserve nonlinear feature learning within each block, motivating its inverse-depth residual scaling.
Weight sharing introduces a separate issue because successive visits can produce correlated residual increments.
Loop-$\mu$P~\cite{wang2026residual} analyzes this recurrence effect and, in our notation, combines $R^{-1}$ loop scaling with $L^{-\frac{1}{2}}$ unique-depth scaling and recurrence-independent block learning rates.
\ours follows the $R^{-1}$ recurrence rule, adopts $L^{-1}$ unique-depth scaling motivated by CompleteP, and adds normalized module-wise and visit-wise residual allocation.
We evaluate this combination in a sandwich topology through both LR transfer and validation loss, then examine the resulting trade-offs across modalities.

\section{Conclusion and future exploration}
\label{sec:conclusion}
\para{Conclusion.} We present \ours{}, a recipe that achieves more stable hyperparameter transfer than prior approaches for Looped Transformers.
\ours uses $\alpha=1$ exponent for both loop axis and unique depth axis.
Our structured module-wise and visit-wise residuals improve the pretraining behavior of looped models. 
Applying \ours{} across vision tasks yields overall gains, and comparing across modalities shows a more favorable trade-off under matched compute in vision pretraining than in language pretraining.

\para{Future exploration.} Due to compute constraints, our current experiments are conducted below the billion-parameter scale. Establishing a solid scaling ladder to larger model sizes will further test the generality of \ours{}. 
In addition, evaluating different model architectures across tasks introduces architectural variation that can confound cross-task comparisons. 
Studying \ours{} within a more unified backbone—and exploring its behavior on multi-step reasoning or post-training regimes—is a promising direction for future work.

\section{Acknowledgment}
\label{sec:acknowledgment}
We thank Matthew Fisher and Kalyan Sunkavalli for supporting compute resources.
We thank Qitao Zhao for his help with E-Rayzer training and data processing.
We thank Nanxuan (Cherry) Zhao, Hao Tan, Yao-Chih Lee, Michael Fischer, Sam Sartor and Nathan Carr for helpful discussions.

\FloatBarrier
\bibliographystyle{assets/plainnat}
\bibliography{references}

\bigskip
\appendix
\setlength{\parskip}{3pt plus 1pt}
\titlespacing*{\paragraph}{0pt}{1ex plus 0.3ex minus 0.2ex}{0.8em}
\captionsetup{skip=4pt}
\setlength{\abovedisplayskip}{4pt plus 2pt minus 1pt}
\setlength{\belowdisplayskip}{4pt plus 2pt minus 1pt}
\setlength{\abovedisplayshortskip}{3pt plus 1pt}
\setlength{\belowdisplayshortskip}{3pt plus 1pt}
\section{Appendix}
\label{app:supplementary}

\subsection{Scalar example for depth scaling}
\label{app:alpha-depth-toy}

\para{Construction.}
We isolate the unique-depth exponent using a scalar Core with shared weights,
omitting the Prelude, Coda, and module/visit factors and setting the fixed
reference-depth constants to one.
For $r=0,\ldots,R-1$ and $\ell=0,\ldots,L-1$, the recurrence is
\begin{equation}
  h_{r,\ell+1}
  =h_{r,\ell}+\frac{(a_\ell b_\ell-1)h_{r,\ell}}{RL^\alpha},
  \qquad h_{r+1,0}=h_{r,L},\quad h_{0,0}=1.
  \label{eq:toy-recurrence}
\end{equation}
The output is $f=h_{R-1,L}$.
The $L$ parameter pairs are shared across visits and initialized at
$a_\ell=b_\ell=1$, so $f=1$ for every $(L,R,\alpha)$ before training.
This identity centering makes the first update transparent; it is not the
random initialization used for a Transformer.

We minimize $\mathcal{L}=\tfrac12(f-2)^2$ using exactly one Adam update,
with LR $\eta L^{\alpha-1}$, $(\beta_1,\beta_2)=(0.9,0.999)$, zero weight decay,
and $\epsilon=0$.
The zero stabilizer is specific to this analytic toy, whose initial gradients
are nonzero; the language models use the positive stabilizer specified in
Section~\ref{sec:hp-transfer}.
Initially, $\partial\mathcal{L}/\partial a_\ell
=\partial\mathcal{L}/\partial b_\ell=-L^{-\alpha}$.
Bias correction therefore gives the same increment for every scalar parameter:
\begin{equation}
  \Delta a_\ell=\Delta b_\ell=\delta,
  \qquad \delta=\eta L^{\alpha-1}.
  \label{eq:toy-adam-update}
\end{equation}

\para{Matched first-order response.}
After the update, each branch coefficient is
\begin{equation}
  a_\ell b_\ell-1
  =\underbrace{\Delta a_\ell+\Delta b_\ell}_{2\delta}
   +\underbrace{\Delta a_\ell\Delta b_\ell}_{\delta^2}.
\end{equation}
The comparator removes only $\delta^2$ from each branch and evaluates the
complete recurrence using the same updated parameters.
It is neither trained independently nor a linearization of the full network
composition.
Writing $N=RL$, the two outputs are exactly
\begin{equation}
  f_{\mathrm{full}}
  =\left(1+\frac{2\eta+\eta^2L^{\alpha-1}}{N}\right)^N,
  \qquad
  f_{\mathrm{lin}}=\left(1+\frac{2\eta}{N}\right)^N.
  \label{eq:toy-exact-outputs}
\end{equation}
Thus the first-order comparator is independent of $\alpha$ at every finite
$L$ and $R$; the comparison isolates the retained within-block interaction.
Its log-output discrepancy is
\begin{equation}
  D=N\log\!\left(1+\frac{\eta^2L^{\alpha-1}}{N+2\eta}\right),
  \qquad
  \frac{D}{\eta^2}\sim L^{\alpha-1}
  \quad (L\to\infty),
  \label{eq:toy-exact-discrepancy}
\end{equation}
for fixed $R,\eta>0$ and $\alpha\in\{0.5,1\}$.
Consequently, $D\to0$ for $\alpha=0.5$, whereas $D\to\eta^2$ for $\alpha=1$.
The normalization by $\eta^2$ removes the leading learning-rate scale;
$D/\eta^2=1$ denotes an output ratio of $\exp(\eta^2)$, not a unit relative
performance gain.

\begin{figure}[!htbp]
  \centering
  \includegraphics[width=0.85\linewidth]{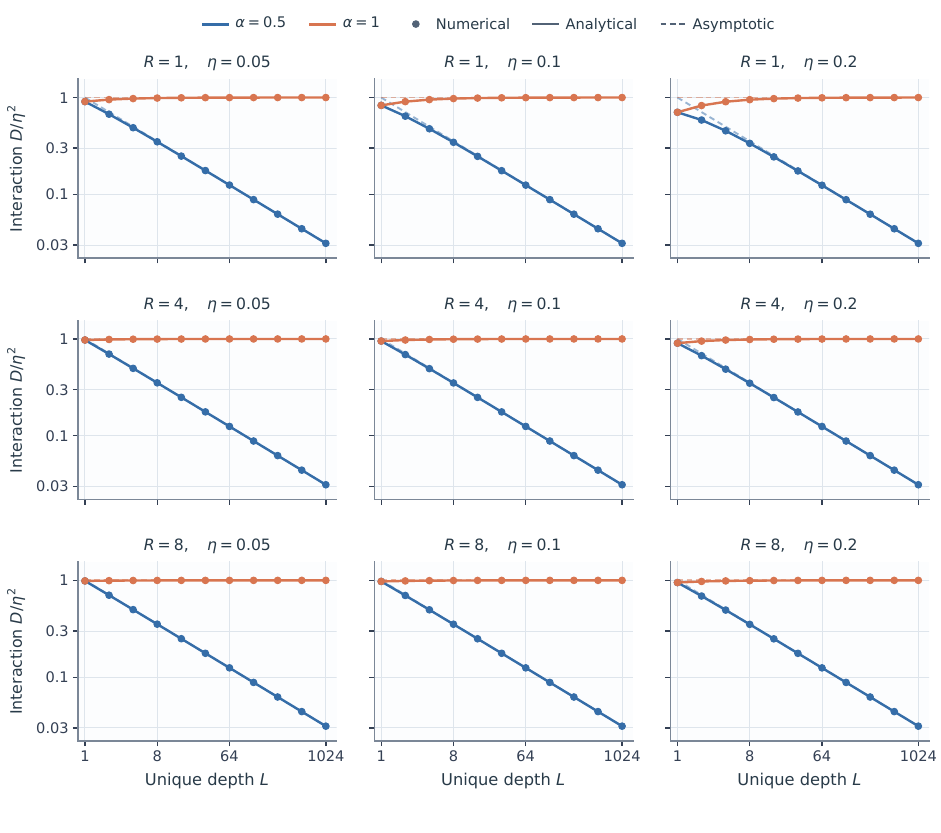}
  \caption{\textbf{Complete scalar depth-scaling sweep.}
  All 198 configurations are shown: 11 unique depths, three recurrence counts,
  two depth exponents, and three base learning rates.
  Rows vary $R$ and columns vary $\eta$; actual Adam LR is $\eta L^{\alpha-1}$.
  Numerical, analytical, and asymptotic quantities are defined in
  Appendix~\ref{app:alpha-depth-toy}; both axes are logarithmic.
  The construction is deterministic, so there are no seed error bars.}
  \label{fig:alpha-depth-toy-full}
\end{figure}

\para{Numerical verification and scope.}
We execute all $11\times3\times2\times3=198$ configurations with
$L=2^k$ for $k=0,\ldots,10$,
$R\in\{1,4,8\}$, $\alpha\in\{0.5,1\}$, and
$\eta\in\{0.05,0.1,0.2\}$.
Each uses explicit $RL$-step unrolling, FP64 CUDA autograd, and one actual Adam
update; the analytical expression does not drive the numerical model.
Figure~\ref{fig:alpha-depth-toy} shows the middle base LR, $\eta=0.1$,
and Figure~\ref{fig:alpha-depth-toy-full} retains every configuration.
At $L=1024$, $R=4$, and $\eta=0.1$, measured $D/\eta^2$ is
$0.0312485$ for $\alpha=0.5$ and $0.9999500$ for $\alpha=1$.
These observations concern within-block parameter interactions after one
update. They do not establish whole-network linearity, multi-step optimization
quality, or a Transformer CE advantage; those require separate empirical evidence.

\subsection{Conditional Analysis of Residual and Optimizer Scaling}
\label{app:theory}

We separate three statements with different assumptions: exact identities for the residual coefficients (Proposition~\ref{prop:scale-invariance}), an asymptotic forward calculation for a coherent scalar Core (Proposition~\ref{prop:depth-scaling}), and an Adam update identity conditional on a prescribed scaling of the entire gradient history (Proposition~\ref{prop:adamw-eps}).
The latter two are illustrative models, not proofs of depth-invariant Transformer optimization or optimal learning rates.

\subsubsection{Coefficient Identities and Correlated Residual Updates}

Recall that the Core branch coefficients are
\begin{equation}
  \gamma_{A,r}=g_{R,L}w_Av_r,\qquad
  \gamma_{F,r}=g_{R,L}w_Fv_r,
\end{equation}
where $w_A=\rho\sqrt{2/(1+\rho^2)}$, $w_F=\sqrt{2/(1+\rho^2)}$, and $v_r=1+\beta\cos\!\left(\pi(r+\frac{1}{2})/R\right)$.

\begin{proposition}[Coefficient Identities Across Modules and Visits]
\label{prop:scale-invariance}
For fixed $\rho>0$, $|\beta|<1$, and integers $R,L\geq1$,
\begin{equation}
  w_A^2+w_F^2=2,\qquad
  \frac{1}{R}\sum_{r=0}^{R-1}v_r=1,\qquad
  \frac{1}{R}\sum_{r=0}^{R-1}v_r^2=
  \begin{cases}
    1,&R=1,\\
    1+\frac{\beta^2}{2},&R\geq2.
  \end{cases}
\end{equation}
Each branch coefficient is bounded above and below by positive, shape-independent multiples of $g_{R,L}$.
These identities concern coefficients, not the moments of hidden states or gradients.
\end{proposition}

\begin{proof}
The module identity follows by adding $2\rho^2/(1+\rho^2)$ and $2/(1+\rho^2)$.
Let $\theta_r=\pi(r+\frac{1}{2})/R$.
Pairing $r$ with $R-1-r$ gives $\sum_r\cos\theta_r=0$.
For $R\geq2$, the geometric-series identity $\sum_r e^{2i\theta_r}=0$ and $\cos^2\theta_r=(1+\cos2\theta_r)/2$ give $\sum_r\cos^2\theta_r=R/2$.
Expanding $v_r$ and $v_r^2$ yields the stated moments; at $R=1$, $v_0=1$.
Finally, $1-|\beta|\leq v_r\leq1+|\beta|$, and both module factors are fixed positive constants.
\end{proof}

\para{Covariance determines cumulative variance.}
Index the branch outputs along the Core trajectory by $i=(b,r,\ell)$, with $b\in\{A,F\}$, and let $X_i$ denote each output after subtracting its expectation.
Writing $\gamma_i=g_{R,L}w_bv_r$, the centered cumulative residual has the exact second moment
\begin{equation}
  \mathbb{E}\left\|\sum_i\gamma_iX_i\right\|^2
  =\sum_{i,j}\gamma_i\gamma_j\,\mathbb{E}\langle X_i,X_j\rangle.
  \label{eq:residual-covariance}
\end{equation}
Under the additional assumption that all distinct branch, layer, and visit contributions are uncorrelated and have the same second moment $\sigma^2$, this reduces to
\begin{equation}
  \mathbb{E}\left\|\sum_i\gamma_iX_i\right\|^2
  =2\left(1+\frac{\beta^2}{2}\mathbf{1}_{R\geq2}\right)
    RL\,g_{R,L}^2\sigma^2.
\end{equation}
Independence across unique layers does not eliminate cross-visit covariance when weights are shared.
For example, in an additive model with $X_{b,r,\ell}=X_{b,\ell}$ at every visit and mutually uncorrelated $(b,\ell)$ contributions of second moment $\sigma^2$, the same quantity is instead
\begin{equation}
  g_{R,L}^2L(w_A^2+w_F^2)\sigma^2
  \left(\sum_r v_r\right)^2
  =2R^2L\,g_{R,L}^2\sigma^2.
\end{equation}
These examples distinguish $RLg_{R,L}^2$ from $R^2Lg_{R,L}^2$ accumulation; neither covariance pattern is assumed to describe the trained Transformer.
Coefficient normalization alone does not fix how output means and covariances change with model shape.

\subsubsection{Asymptotic Forward Scaling Under Coherent Drift}

CompleteP~\cite{dey2025completep} adopts $\alpha_L=1$, while Depth-$\mu$P~\cite{yang2024tensor} and Loop-$\mu$P~\cite{wang2026residual} use $\alpha_L=\frac{1}{2}$.
To illustrate one motivation for stronger depth scaling, we explicitly assume a common positive drift in a scalar linear Core.
Weight sharing by itself does not establish this assumption.
We use the normalized $g_{R,L}$ from Eq.~\eqref{eq:global-residual}, with $\alpha_R=1$, and omit the module and visit factors in this example.

\begin{proposition}[Asymptotic Amplification in a Coherent Scalar Core]
\label{prop:depth-scaling}
Fix positive integers $R,R_0,L_0$ and $\mu>0$, and consider
\begin{equation}
  x_{r,\ell+1}=(1+\mu g_{R,L})x_{r,\ell},\qquad
  x_{r+1,0}=x_{r,L},\qquad
  g_{R,L}=\frac{R_0}{R}m_L^{-\alpha_L},
\end{equation}
with $x_{0,0}>0$ and $m_L=L/L_0$.
Let $c=\mu R_0L_0$ and $\mathcal{A}_{R,L}=x_{R,0}/x_{0,0}$.
For $\alpha_L=1$,
\begin{equation}
  \mathcal{A}_{R,L}=\left(1+\frac{c}{RL}\right)^{RL}
  \leq e^c,\qquad
  \log\mathcal{A}_{R,L}
  =c-\frac{c^2}{2RL}+O\!\left((RL)^{-2}\right).
  \label{eq:coherent-depth-limit}
\end{equation}
Thus $\mathcal{A}_{R,L}\to e^c$ as $L\to\infty$, but the amplification is not exactly invariant at finite depth.
For $\alpha_L=\frac{1}{2}$ under the same positive-drift assumption,
\begin{equation}
  \log\mathcal{A}_{R,L}
  =c\,m_L^{\frac{1}{2}}+O(1)
  \qquad (L\to\infty,\ R\text{ fixed}).
\end{equation}
\end{proposition}

\begin{proof}
The scalar recurrence gives $\mathcal{A}_{R,L}=(1+\mu g_{R,L})^{RL}$ exactly.
For $\alpha_L=1$, $\mu g_{R,L}=c/(RL)$; the bound follows from $\log(1+u)\leq u$ for $u\geq0$, and the expansion follows from the Taylor series of $\log(1+u)$.
For $\alpha_L=\frac{1}{2}$, the first Taylor term is $RL\mu g_{R,L}=c\,m_L^{\frac{1}{2}}$, the second is $O(1)$, and the remaining terms are $O(m_L^{-\frac{1}{2}})$ at fixed $R$.
\end{proof}

For example, $c=1$ and $R=1$ give amplification factors $2$ and $9/4$ at $L=1$ and $L=2$, respectively, despite the common limit $e$.
The result motivates controlling a persistent positive drift; it does not assert that Transformer residuals follow this scalar dynamics.
In particular, bounded forward activations do not establish bounds or invariance for backpropagated Jacobians, parameter updates, or optimal learning rates.

\subsubsection{Conditional Scaling of the Adam Stabilizer}

The depth factor in the AdamW stabilizer can be motivated separately by the algebra of Adam~\cite{kingma2014adam}.
We hold width fixed and suppress its constant multiplier below.
The following statement concerns the adaptive gradient term of AdamW, excluding decoupled weight decay.

\begin{proposition}[Adam Homogeneity for a Prescribed Gradient History]
\label{prop:adamw-eps}
For a fixed parameter coordinate, suppose the gradients actually supplied to Adam satisfy
\begin{equation}
  G_t^{(L)}=m_L^{-1}G_t^{\mathrm{ref}}
  \quad\text{for every step }t.
  \label{eq:conditional-gradient-history}
\end{equation}
Here $\mathrm{ref}$ denotes a prescribed reference-depth gradient sequence.
Assume zero-initialized moments, common $(\beta_1,\beta_2)$ and LR schedule $\eta_t$, and $\epsilon_{\rm base}>0$.
Then the bias-corrected moments obey
\begin{equation}
  \widehat{M}_t^{(L)}=m_L^{-1}\widehat{M}_t^{\mathrm{ref}},
  \qquad
  \widehat{V}_t^{(L)}=m_L^{-2}\widehat{V}_t^{\mathrm{ref}}.
\end{equation}
With $\epsilon_{\rm Core}^{(L)}=\epsilon_{\rm base}m_L^{-1}$, the adaptive update is exactly
\begin{equation}
  \Delta W_t^{\mathrm{adapt},(L)}
  =-\eta_t\frac{m_L^{-1}\widehat{M}_t^{\mathrm{ref}}}
    {m_L^{-1}\sqrt{\widehat{V}_t^{\mathrm{ref}}}
      +m_L^{-1}\epsilon_{\rm base}}
  =-\eta_t\frac{\widehat{M}_t^{\mathrm{ref}}}
    {\sqrt{\widehat{V}_t^{\mathrm{ref}}}+\epsilon_{\rm base}}.
  \label{eq:conditional-adam-homogeneity}
\end{equation}
\end{proposition}

\begin{proof}
Adam's first moment is an exponential average of the gradient history and its second moment is an exponential average of its elementwise square.
Equation~\eqref{eq:conditional-gradient-history} therefore scales these averages by $m_L^{-1}$ and $m_L^{-2}$, respectively.
The common bias-correction factors preserve these identities; substitution yields Eq.~\eqref{eq:conditional-adam-homogeneity}.
\end{proof}

The proportional gradient histories are an assumption, not a consequence of bounded activations.
For a shared branch parameter, the chain rule gives
\begin{equation}
  G_{b,\ell}=g_{R,L}w_b\sum_r v_r\widetilde{G}_{b,r,\ell},
\end{equation}
where $\widetilde{G}_{b,r,\ell}$ includes the local parameter derivative and all downstream Jacobian factors.
Even if these contributions were uniformly bounded, at $\alpha_R=\alpha_L=1$ the coefficient sum would imply only $G_{b,\ell}=O(m_L^{-1})$; cancellations can make it smaller, and its direction can vary with depth.
This bound does not establish Eq.~\eqref{eq:conditional-gradient-history}, including after any gradient clipping.

Under the proposition's prescribed history, keeping $\epsilon_{\rm Core}$ fixed instead changes the adaptive denominator to $\sqrt{\widehat{V}_t^{\mathrm{ref}}}+\epsilon_{\rm base}m_L$ after canceling the gradient scale.
This explains the role of matching the stabilizer to an assumed gradient scale.
It does not determine the optimal LR, preservation of feature learning, or equivalence of training trajectories across depths; HP transfer is evaluated empirically in Section~\ref{sec:experiments_language}.

\subsection{Language Model Architecture}
\label{app:language-architecture}

Table~\ref{tab:language-architecture} summarizes the architectural settings shared by our language-modeling experiments.
The models use Llama-style pre-norm decoder blocks~\cite{touvron2023llama}.
The sandwich topology and residual coefficients follow Section~\ref{sec:topology}.

\begin{table}[!htbp]
  \centering
  \small
  \setlength{\tabcolsep}{6pt}
  \renewcommand{\arraystretch}{1.18}
  \caption{Base architecture for the language-modeling experiments.}
  \label{tab:language-architecture}
  \begin{tabularx}{\linewidth}{@{}>{\raggedright\arraybackslash}p{0.27\linewidth}
    >{\raggedright\arraybackslash}X@{}}
    \toprule
    Component & Setting \\
    \midrule
    Model type & Decoder-only, pre-norm Transformer \\
    Hidden dimension & $768$ \\
    Attention & Full-sequence causal self-attention \\
    Attention heads & $12$ query, $12$ key, and $12$ value heads \\
    Head dimension & $64$ \\
    Feed-forward network & SwiGLU with intermediate dimension $2048$ \\
    Normalization & Two RMSNorm modules per block and one final RMSNorm before the language-model head; learned gains and stabilizer $10^{-6}$ \\
    Block order & RMSNorm $\to$ Attention $\to$ residual addition $\to$\newline
      RMSNorm $\to$ SwiGLU $\to$ residual addition \\
    Positional encoding & RoPE with base $1{,}000{,}000$, applied to all head dimensions \\
    Context length & $4096$ tokens \\
    Vocabulary & $32{,}000$ tokens, Llama tokenizer \\
    Embedding / LM head & Untied weights; each matrix has shape $32{,}000\times768$ \\
    Bias & No bias in attention or feed-forward linear projections \\
    Other settings & Dropout, query/key normalization, and attention output gates are disabled \\
    \bottomrule
  \end{tabularx}
\end{table}

\subsection{Recurrence LR Scaling Analysis}
\label{app:recurrence-lr-analysis}

\para{How to transfer HPs with recurrence $R$.}
When the same Core is visited $R$ times with shared weights, increasing $R$ changes both the effective update depth and representation alignment, raising a basic question for hyperparameter transfer: how should the optimal learning rate scale with $R$?
Motivated by prior work on scaling and HP transfer~\cite{dey2025completep,yang2024tensor,wang2026residual}, we compare two candidate rules: recurrence-independent optimal block learning rates and optimal block learning rates that decrease as $R^{-\frac{1}{2}}$.
Table~\ref{tab:parameterization} pairs the former with $R^{-1}$ residual scaling ($\alpha_R=1$).
We hold this residual rule fixed and test whether its LR-transfer prescription is supported by the measured curves.

\begin{figure}[!htbp]
  \centering
  \begin{subfigure}[t]{0.43\textwidth}
    \includegraphics[width=\linewidth]{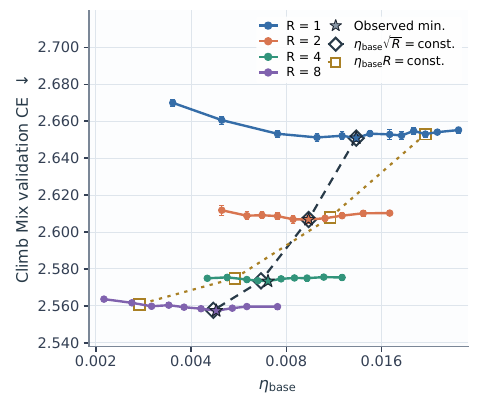}
    \caption{Optimizer input $\eta_{\rm base}$.}
    \label{fig:r-axis-input-lr}
  \end{subfigure}\hspace{0.04\textwidth}
  \begin{subfigure}[t]{0.43\textwidth}
    \includegraphics[width=\linewidth]{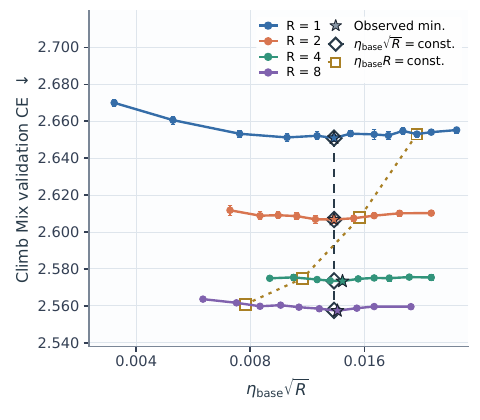}
    \caption{LR coordinates shifted to $\eta_{\rm base}\sqrt{R}$.}
    \label{fig:r-axis-block-lr}
  \end{subfigure}
  \caption{\textbf{Comparing recurrence HP transfer.}
    The same Climb Mix~\cite{diao2025climb} validation losses are shown in two LR coordinates.
    The measured minima shift toward smaller optimizer inputs as $R$ grows, following $\eta_{\rm base}\sqrt{R}=\mathrm{const.}$ more closely than $\eta_{\rm base}R=\mathrm{const.}$.
    Multiplying the input LR by $\sqrt{R}$ approximately aligns these minima, supporting $\alpha_\eta=1$ and a common block LR.
    Each candidate rule uses its own fitted constant.}
  \label{fig:r-axis-lr-coordinate}
\end{figure}

\para{Fitting the empirical LR exponent.}
We first scan the optimizer input $\eta_{\rm base}$ independently at each recurrence count $R$ to obtain validation-loss curves.
In this experiment, the optimizer multiplies the input $\eta_{\rm base}$ by $\sqrt{R}$ to set the block learning rates, up to fixed parameter-group multipliers.
Figure~\ref{fig:r-axis-lr-coordinate} shows the same scans in two coordinates: the optimizer input $\eta_{\rm base}$ and the resulting block-LR coordinate $\eta_{\rm base}\sqrt{R}$.
For comparison with the LR entries in Table~\ref{tab:parameterization}, the relevant quantity is the latter: keeping it fixed preserves the block learning rates across recurrence counts.
We hypothesize that multiplying recurrence by a fixed factor changes the optimal block LR by a fixed ratio, independent of the starting recurrence count.
We express this power law as
\begin{equation}
  \eta_{\rm base}^{\star}(R)\sqrt{R}
  \propto R^{\alpha_\eta-1},
  \label{eq:recurrence-lr-power-law}
\end{equation}
where $\eta_{\rm base}^{\star}(R)$ denotes the input LR that minimizes validation loss at recurrence $R$.
This convention makes $\alpha_\eta=1$ the rule that preserves block LR, while $\alpha_\eta=\frac{1}{2}$ reduces it as $R^{-\frac{1}{2}}$.
We estimate $\alpha_\eta$ from the measured loss curves; it describes LR transfer, while $\alpha_R$ controls the residual scaling held fixed in this experiment.

To relate these hypotheses to the scanned input, Eq.~\eqref{eq:recurrence-lr-power-law} gives
\begin{equation}
  \eta_{\rm base}^{\star}(R)
  \propto R^{\alpha_\eta-\frac{3}{2}}.
  \label{eq:recurrence-lr-coordinate-scaling}
\end{equation}
If $\alpha_\eta=1$, the minima move left as $R^{-\frac{1}{2}}$ on the input-LR axis.
Multiplying by $\sqrt{R}$ cancels this movement, so the minima should align in the block-LR coordinate:
\begin{equation}
  \eta_{\rm base}^{\star}(R)\sqrt{R}
  \propto R^{-\frac{1}{2}}\sqrt{R}
  = R^0.
  \label{eq:recurrence-lr-cancellation}
\end{equation}
If $\alpha_\eta=\frac{1}{2}$, the input minima move left more rapidly, as $R^{-1}$, and still shift left as $R^{-\frac{1}{2}}$ after conversion.
We therefore construct nonuniform LR grids that cover both predicted locations and their neighborhoods, sampling both sides of the potential minima.
This design lets each curve locate its own minimum before either transfer rule is assessed.

\para{What the curves reveal.}
We compare two candidate rules: $\eta_{\rm base}\sqrt{R}=\mathrm{const.}$ ($\alpha_\eta=1$) and $\eta_{\rm base}R=\mathrm{const.}$ ($\alpha_\eta=\frac{1}{2}$).
For each rule, we fit a single shared constant across all recurrence counts.
Each trial constant selects an LR for every $R$; interpolating the measured loss curves gives the corresponding validation losses.
We vary the constant to minimize the total excess loss above each curve's observed minimum across recurrence counts.
Connecting the selected points gives the candidate paths in Figure~\ref{fig:r-axis-lr-coordinate}.
In Fig.~\ref{fig:r-axis-input-lr}, the measured minima shift toward smaller input LRs as recurrence increases, following the $R^{-\frac{1}{2}}$ input rule more closely than the faster $R^{-1}$ rule.
Figure~\ref{fig:r-axis-block-lr} makes the implication for transfer visible: the same minima approximately align after multiplying the input LR by $\sqrt{R}$.
Even with both constants optimized, the $\alpha_\eta=1$ path stays close to these valleys, while the $\alpha_\eta=\frac{1}{2}$ path continues to move left and lies farther from the measured minima.
Thus, the curves favor $\alpha_\eta=1$, supporting the recurrence-independent block LRs paired with $\alpha_R=1$ in Table~\ref{tab:parameterization}.

\para{Quantifying the apparent alignment.}
The comparison in Fig.~\ref{fig:r-axis-lr-coordinate} tests two fixed exponents; we now let $\alpha_\eta$ vary to estimate the scaling preferred by the data.
Since the best sampled LR can move with grid spacing or seed variability, we use the surrounding losses to locate each valley.
Working in log LR makes equal LR ratios correspond to equal horizontal shifts, matching the multiplicative scaling laws we wish to test.
A quadratic then provides a simple model of a smooth valley:
\begin{equation}
  \overline{\mathrm{CE}}_R(\eta_{\rm base})
  \approx A_R+B_R
  \left[\log\frac{\eta_{\rm base}}{\eta_{\rm base}^{\star}(R)}\right]^2,
  \qquad B_R\geq0.
  \label{eq:recurrence-lr-fit}
\end{equation}
Here $\overline{\mathrm{CE}}_R$ is the mean validation CE across seeds.
The squared log-ratio measures multiplicative distance from the fitted optimum $\eta_{\rm base}^{\star}(R)$: it vanishes at that LR and grows on either side.
Each curve has its own valley height $A_R$ and steepness $B_R$, while Eq.~\eqref{eq:recurrence-lr-power-law} links the valley locations through the shared parameter $\alpha_\eta$.
We jointly fit these curves across the scanned range by minimizing squared error to the measured losses.
This estimates how the optimal LR shifts with $R$ while allowing the loss level and LR sensitivity to change independently.

\begin{figure}[!htbp]
  \centering
  \includegraphics[width=0.44\textwidth]{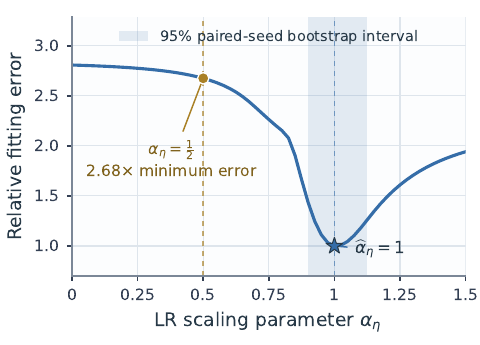}
  \caption{\textbf{Estimating the LR scaling.}
    The loss-curve fit favors $\alpha_\eta=1$, corresponding to a common block LR across recurrence counts.
    The shaded bootstrap interval includes $1$ and excludes $\frac{1}{2}$, whose fit gives a larger error.
    Errors are normalized by their minimum.}
  \label{fig:r-axis-fit-profile}
\end{figure}

Fig.~\ref{fig:r-axis-fit-profile} shows how the best fitting error changes as $\alpha_\eta$ varies, reoptimizing the shared LR constant and each curve's height and steepness at every exponent.
The error is normalized by its minimum, so the curve directly compares how well different exponents explain the same measured losses.
The fitted value is $\widehat{\alpha}_\eta=1$, and the $95\%$ paired-seed bootstrap interval includes $1$ but excludes $\frac{1}{2}$.
Thus, allowing the exponent to vary recovers the preference seen in the two coordinate views: the optimal input LR offsets the optimizer's $\sqrt{R}$ multiplier, leaving the optimal block LR approximately unchanged.
The precision depends on the curve approximation; fits restricted to smaller neighborhoods of the minima are less precise.
Together, the visual comparison and quantitative fit support the constant block LR used with $\alpha_R=1$ in our recipe.
This conclusion concerns LR transfer under the fixed $R^{-1}$ residual rule and the historical non-block LR policy; it does not estimate the residual exponent $\alpha_R$.

\subsection{Unique-Depth and Mixed-Shape Transfer Details}
\label{app:depth-transfer-details}

\para{Coordinates and model shapes.}
Figures~\ref{fig:depth-transfer} and~\ref{fig:mixed-transfer} evaluate Climb Mix~\cite{diao2025climb} validation CE after the same $15.27\text{B}$ training tokens for each model.
The displayed \ours grid is $\{0.00665,0.00940452,0.01330,0.01880904,0.02660\}$, and the Loop-$\mu$P grid is $\{0.000892,0.001261,0.001784,0.002523,0.003568\}$.
Figure~\ref{fig:depth-transfer} fixes $R=4$ and varies $L\in\{2,4,8,12\}$; Figure~\ref{fig:mixed-transfer} compares the same $(R,L)=(4,4)$ reference with $(8,8)$ and $(8,12)$.

\para{Measured minima and transfer.}
All displayed \ours curves select LR $\eta_{\rm base}=0.01330$ on the sampled grid.
Table~\ref{tab:depth-transfer-minima} compares the mean CE at that shared LR against each matched Loop-$\mu$P curve's lowest measured mean, which supplies the horizontal reference in Figures~\ref{fig:depth-transfer-b-axis} and~\ref{fig:depth-transfer-c-axis}.

\begin{table}[!htbp]
  \centering
  \small
  \caption{\textbf{Observed minima in the depth-transfer sweeps.}
    \ours uses the same $\eta_{\rm base}=0.01330$ in every row; Loop-$\mu$P uses its best tested LR separately for each shape.
    }
  \label{tab:depth-transfer-minima}
  \begin{tabular}{rrrrrr}
    \toprule
    $R$ & $L$ & $n_{\mathrm{ours}}$ & $n_{\mathrm{Loop}}$ & \ours CE & Loop-$\mu$P CE  \\
    \midrule
    4 & 2  & 5 & 5 & 2.6447 & 2.6720  \\
    4 & 4  & 5 & 5 & 2.5680 & 2.5890  \\
    4 & 8  & 5 & 5 & 2.4906 & 2.5070  \\
    4 & 12 & 5 & 5 & 2.4435 & 2.4577  \\
    8 & 8  & 5 & 5 & 2.4702 & 2.4870  \\
    8 & 12 & 5 & 5 & 2.4250 & 2.4395  \\
    \bottomrule
  \end{tabular}
\end{table}

\subsection{Vision Training and Evaluation Details}
\label{app:vision-protocols}

\para{Task-specific residual allocation.}
The E-RayZer~\cite{zhao2026erayzer} and PixelDiT~\cite{yu2026pixeldit} recurrence experiments retain their native architectures, optimizers, and gates.
Their module allocation $\rho=2$, and visit allocation $\beta=-0.5$, with unscaled Prelude/Coda residuals and uniform native task LRs.

\para{E-RayZer training and evaluation.}
Both encoders have hidden width 768 and 12 attention heads, retaining the native alternation between within-frame and cross-frame attention.
Training uses AdamW~\cite{loshchilov2019decoupled} with LR $4\times10^{-4}$, $(\beta_1,\beta_2)=(0.9,0.95)$, and matrix weight decay $0.05$; vector parameters have zero weight decay.
The schedule has 3,000 warmup iterations and a cosine horizon of 156,210 iterations, with evaluation at iteration 152,000.
The training loss combines image MSE and a VGG-based perceptual term~\cite{simonyan2015vgg,johnson2016perceptual} weighted by $0.2$.
The held-out DL3DV~\cite{ling2024dl3dv} benchmark contains 140 scenes and 700 target images, using five reference and five target views per scene with a fixed frame-distance protocol.
PSNR is aggregated over target views and then scenes.

\para{PixelDiT training and evaluation.}
The patch Transformer has hidden width 1,152 and patch size 16; the four pixel-level blocks remain independent.
REPA~\cite{yu2025repa} aligns the output of the eighth Prelude block to frozen DINOv2~\cite{oquab2023dinov2} features, once per forward pass, with loss weight $0.5$.
AdamW~\cite{loshchilov2019decoupled} uses LR $10^{-4}$, $(\beta_1,\beta_2)=(0.9,0.999)$, zero weight decay, and gradient clipping at one; EMA decay is $0.9999$.
The endpoints in Table~\ref{tab:benchmark-400320} use 400,320 updates with global batch 256.
Evaluation uses the ADM metric protocol~\cite{dhariwal2021diffusion} with 50K EMA samples, 100 sampling steps, guidance scale $3.25$ over the interval $[0.1,1]$, and time shift one.

\para{Text-to-image architecture and data.}
MiniT2I-B/32~\cite{minit2i2026} uses hidden width 768, 12 attention heads, SwiGLU~\cite{shazeer2020glu} width 2,048, and two text-preamble blocks, with a maximum of 256 text tokens.
The CC12M~\cite{changpinyo2021cc12m} subset contains 10,952,154 training examples and two disjoint sets of 8,192 examples each for validation and held-out reporting.
The SFT mixture contains 58,859 BLIP3o-60K~\cite{chen2025blip3o}, 19,024 DALL-E 3~\cite{betker2023dalle3,opendatasets2024dalle3}, and 41,356 ShareGPT-4o~\cite{chen2025sharegpt4o} examples, sampled with relative source weights $0.06:0.016:0.04$ following the native MiniT2I recipe.

\para{Text-to-image training and evaluation.}
Table~\ref{tab:pretrain-sft} reports 250K pretraining updates followed by an additional 40K SFT updates, with global batch 1024.
The denoiser predicts the clean image and is trained with the corresponding velocity MSE at the final output only.
AdamW~\cite{loshchilov2019decoupled} uses $(\beta_1,\beta_2)=(0.9,0.95)$, zero weight decay, and no gradient clipping; EMA decay is $0.99995$.
Pretraining uses a 5K-step warmup followed by a constant LR. The unscaled models use LR $4\times10^{-4}$; the residual-scaled variant uses $1.6\times10^{-3}$, $\rho=1$, and $\beta=0.5$.
Pretraining uses BF16, whereas SFT uses FP32 and retains the optimizer and EMA states without a new warmup.
GenEval~\cite{ghosh2023geneval} and DPG~\cite{hu2024ella} use EMA weights, classifier-free guidance~\cite{ho2022classifierfree} with scale six, Euler sampling with 100 steps, and paired FP32 evaluation.
Post-SFT FID~\cite{heusel2017gans} uses 30K COCO~\cite{lin2014coco} samples, reported separately for EMA weight.

\clearpage
\newgeometry{left=18bp,right=18bp,top=12bp,bottom=18bp}
\thispagestyle{empty}
\noindent\begin{minipage}{\textwidth}
  \centering
  \includegraphics[
    trim=18bp 11.666667bp 18bp 11.666667bp,clip,
    width=\linewidth,height=\dimexpr\textheight-30pt\relax,keepaspectratio
  ]{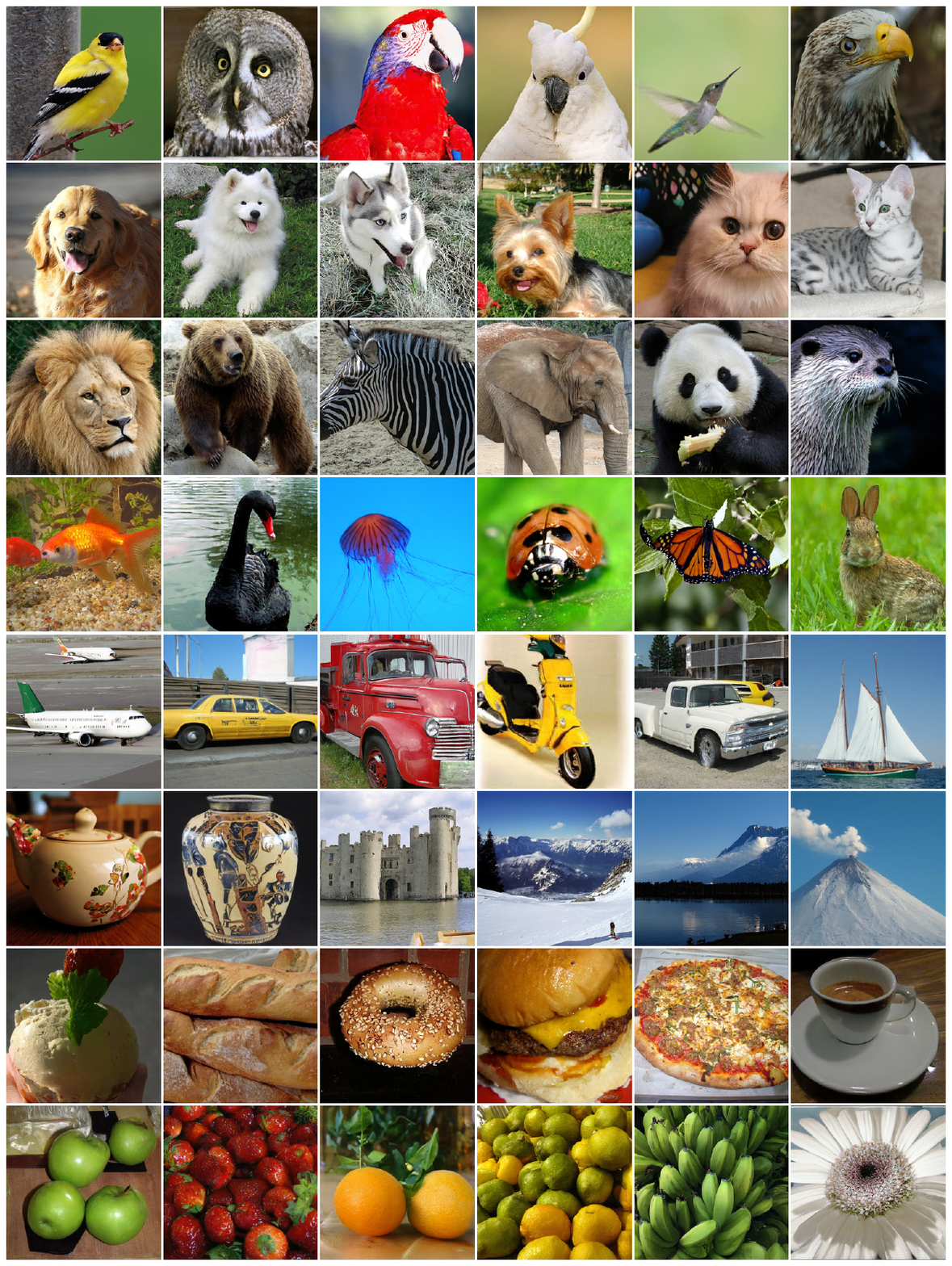}
  \captionsetup{justification=centering,singlelinecheck=false,skip=8pt}
  \captionof{figure}{Images generated by our looped PixelDiT~\cite{yu2026pixeldit}, ImageNet~\cite{deng2009imagenet} $256 \times 256$.}
  \label{fig:pixeldit-loop-gallery}
\end{minipage}
\restoregeometry

\clearpage
\newgeometry{left=18bp,right=18bp,top=12bp,bottom=18bp}
\thispagestyle{empty}
\noindent\begin{minipage}{\textwidth}
  \centering
  \includegraphics[
    trim=18bp 11.666667bp 18bp 11.666667bp,clip,
    width=\linewidth,height=\dimexpr\textheight-30pt\relax,keepaspectratio
  ]{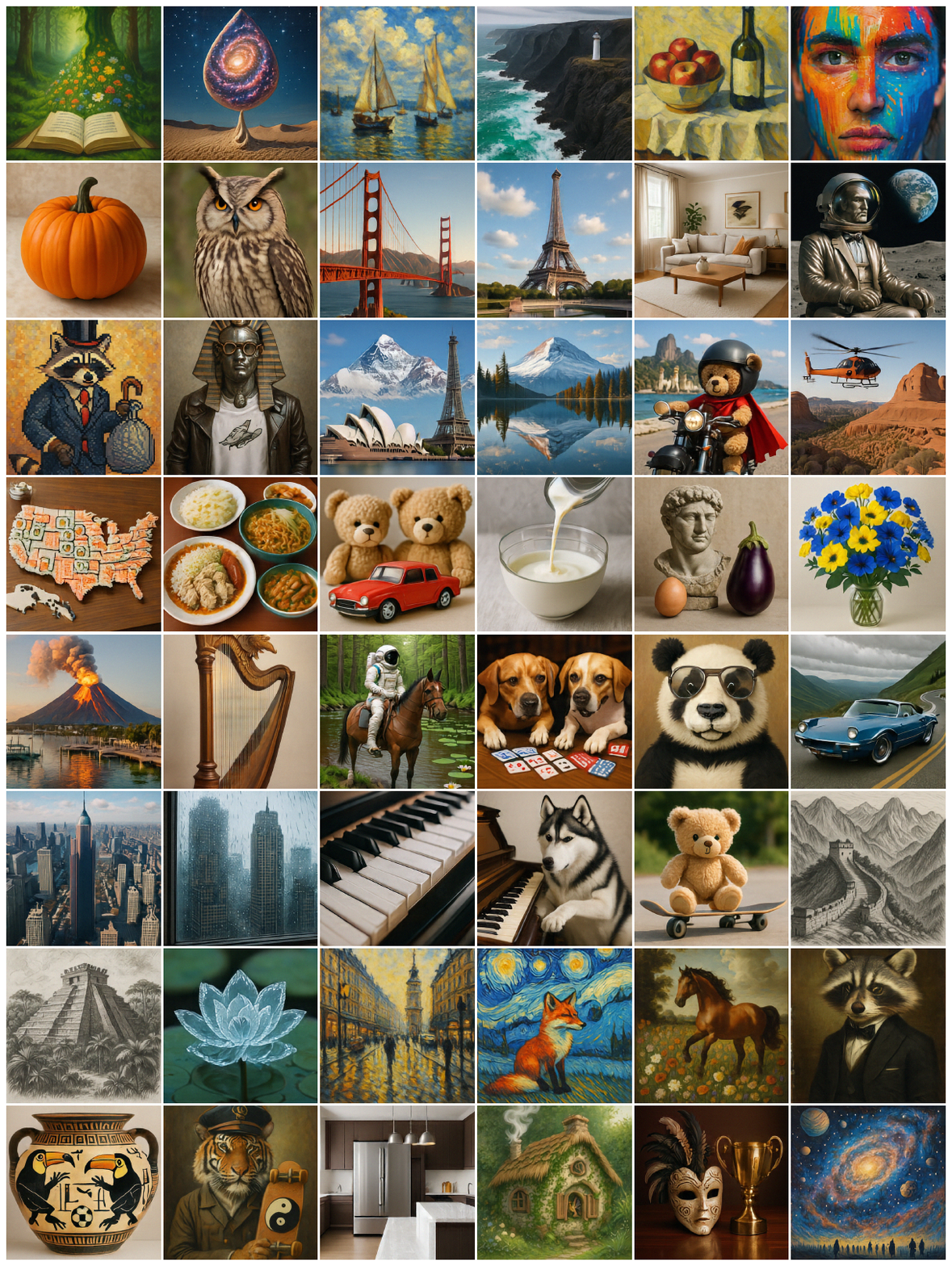}
  \captionsetup{justification=centering,singlelinecheck=false,skip=8pt}
  \captionof{figure}{Images generated by our looped MiniT2I~\cite{minit2i2026}.}
  \label{fig:minit2i-loop-gallery}
\end{minipage}
\restoregeometry

\clearpage
\newgeometry{left=30bp,right=30bp,top=30bp,bottom=30bp}
\thispagestyle{empty}
{\large\sffamily\bfseries Prompts for Figure~\ref{fig:minit2i-loop-gallery}\par}
\smallskip
{\small Numbered in raster order: left to right across each row, then top to bottom.\par}
\medskip
\noindent\begin{minipage}[t]{0.48\textwidth}
  \raggedright\fontsize{9}{11}\selectfont
  \setlength{\parindent}{0pt}\setlength{\parskip}{4pt}
  \textbf{1.} An open book lies at the threshold of a mystical forest, its pages seamlessly merging with the verdant landscape, where vibrant flowers bloom directly from the text, weaving a tangible tapestry of stories.\par
  \textbf{2.} A cosmic teardrop carved from the desert sand cradles a swirling galaxy, defying gravity as it floats serenely against a starlit sky.\par
  \textbf{3.} Impressionist oil painting of sailboats on a river under a bright cloudy sky.\par
  \textbf{4.} An aerial photograph of a rugged coastline, with turquoise waves crashing against black volcanic cliffs. A solitary lighthouse stands on the edge of the promontory, its white structure a stark contrast to the dark rock.\par
  \textbf{5.} A Post-Impressionist still life of a bowl of apples and a wine bottle on a rumpled tablecloth. The objects are simplified into basic geometric forms, with a flattened perspective and strong, visible brushwork emphasizing structure over realism.\par
  \textbf{6.} Detailed, close-up photograph of a person's face, artistically covered in a vibrant and expressive layer of multi-colored paint. The camera is focused sharply on the subject's left eye, creating a shallow depth of field that softly blurs the rest of the face and the background. The paint is applied in thick, bold strokes and dabs, creating a vivid rainbow portrait that celebrates color, creativity, and self-expression.\par
  \textbf{7.} a pumpkin\par
  \textbf{8.} an owl\par
  \textbf{9.} a photo of san francisco's golden gate bridge\par
  \textbf{10.} the Eiffel Tower\par
  \textbf{11.} a living room\par
  \textbf{12.} a statue of Abraham Lincoln wearing an opaque and shiny astronaut's helmet. The statue sits on the moon, with the planet Earth in the sky.\par
  \textbf{13.} A raccoon wearing formal clothes, wearing a tophat and holding a cane. The raccoon is holding a garbage bag. Oil painting in the style of pixel art.\par
  \textbf{14.} A portrait of a metal statue of a pharaoh wearing steampunk glasses and a leather jacket over a white t-shirt that has a drawing of a space shuttle on it.\par
  \textbf{15.} the Sydney Opera House with the Eiffel tower sitting on the right, and Mount Everest rising above\par
  \textbf{16.} Snow mountain and tree reflection in the lake\par
  \textbf{17.} A teddy bear wearing a motorcycle helmet and cape is riding a motorcycle in Rio de Janeiro with Dois Irmãos in the background\par
  \textbf{18.} A helicopter flies over the Arches National Park.\par
  \textbf{19.} A map of the United States made out of sushi on the table.\par
  \textbf{20.} A table full of food. There is a plate of chicken rice, a bowl of bak chor mee, and a bowl of laksa.\par
  \textbf{21.} a toy car in front of a teddy bear\par
  \textbf{22.} milk pouring from a glass into a bowl\par
  \textbf{23.} a stone bust next to an egg and an eggplant\par
  \textbf{24.} a bundle of blue and yellow flowers in a vase\par
\end{minipage}\hfill%
\noindent\begin{minipage}[t]{0.48\textwidth}
  \raggedright\fontsize{9}{11}\selectfont
  \setlength{\parindent}{0pt}\setlength{\parskip}{4pt}
  \textbf{25.} a volcano exploding next to a marina\par
  \textbf{26.} a harp with a carved eagle figure at the top\par
  \textbf{27.} A photo of an astronaut riding a horse in the forest. There is a river in front of them with water lilies.\par
  \textbf{28.} Dogs playing poker\par
  \textbf{29.} a panda bear with aviator glasses on its head\par
  \textbf{30.} Three-quarters front view of a blue 1977 Corvette coming around a curve in a mountain road and looking over a green valley on a cloudy day.\par
  \textbf{31.} an overhead view of the Empire State Building\par
  \textbf{32.} tall buildings seen through a window with rain on it\par
  \textbf{33.} a close-up of the keys of a piano\par
  \textbf{34.} Siberian husky playing the piano.\par
  \textbf{35.} a teddy bear on a skateboard\par
  \textbf{36.} A section of the Great Wall in the mountains. detailed charcoal sketch.\par
  \textbf{37.} A Mesoamerican pyramid surrounded by jungle. detailed charcoal sketch.\par
  \textbf{38.} A photo of a lotus flower made of water.\par
  \textbf{39.} a painting of street in Paris\par
  \textbf{40.} a painting of a fox in the style of starry night\par
  \textbf{41.} a dutch baroque painting of a horse in a field of flowers\par
  \textbf{42.} portrait of a well-dressed raccoon, oil painting in the style of Rembrandt\par
  \textbf{43.} A photo of an Athenian vase with a painting of toucans playing soccer in the style of Egyptian hieroglyphics\par
  \textbf{44.} Portrait of a tiger wearing a train conductor's hat and holding a skateboard that has a yin-yang symbol on it. oil painting\par
  \textbf{45.} a spacious kitchen featuring a large stainless steel refrigerator with a double-door design. The refrigerator stands next to sleek, dark wooden cabinets that reach up to the ceiling. In front of the refrigerator, there is a kitchen island with a white marble countertop, and hanging above are three modern pendant lights with a brushed metal finish.\par
  \textbf{46.} A quaint fairy cottage nestled in a lush garden, with delicate smoke wisps rising from its stone chimney. The cottage features a thatched roof and walls covered in climbing ivy. A curious squirrel peers out from a small, round window, framed by wooden shutters.\par
  \textbf{47.} A traditional Venetian mask with intricate designs and feather embellishments is placed on a polished wooden table. Next to the mask, there sits a substantial golden trophy, shining with reflected light, its surface etched with small, detailed engravings. The table and its prestigious items are set against a deep sepia-toned backdrop that enhances the objects' visual appeal.\par
  \textbf{48.} A striking image depicting an imagined distant galaxy teeming with life, inspired by the work of the renowned artist Caspar David Friedrich, is showcased in high quality on Artstation. The matte painting presents a vibrant scene filled with ethereal colors and pulsating with otherworldly energy. As viewers delve into the scene, they encounter intricate details of celestial bodies and the silhouettes of people populating this fantasy cosmos.\par
\end{minipage}
\restoregeometry

\clearpage
\newgeometry{left=18bp,right=18bp,top=12bp,bottom=12bp}
\thispagestyle{empty}
\noindent\makebox[\textwidth][c]{%
\begin{minipage}{420bp}
  \centering
  \includegraphics[width=\linewidth,height=\dimexpr\textheight-48pt\relax,keepaspectratio]{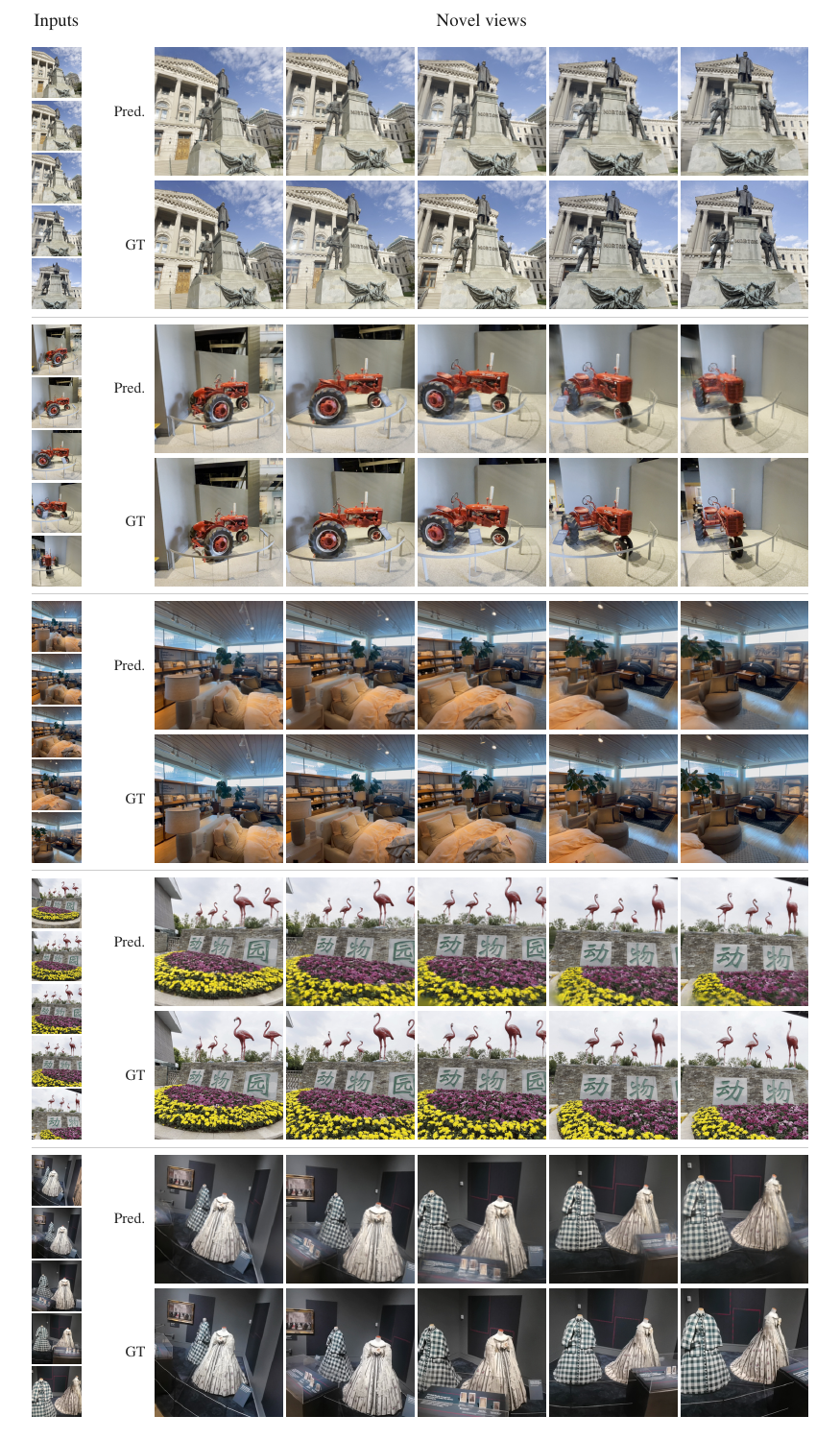}
  \captionsetup{justification=raggedright,singlelinecheck=false,skip=6pt}
  \captionof{figure}{Novel-view synthesis with our looped E-RayZer~\cite{zhao2026erayzer} on the DL3DV~\cite{ling2024dl3dv} test split. Each scene shows five input views (left), five predicted novel views (upper right), and the corresponding ground truth (lower right).}
  \label{fig:erayzer-loop-gallery}
\end{minipage}}
\restoregeometry

\end{document}